\def\year{2021}\relax
\documentclass[letterpaper]{article} 
\usepackage{aaai21}  
\usepackage{times}  
\usepackage{helvet} 
\usepackage{courier}  
\usepackage[hyphens]{url}  
\usepackage{graphicx} 
\usepackage{natbib}  
\usepackage{caption} 
\usepackage{microtype}
\usepackage{soul}
\usepackage{url}
\usepackage[utf8]{inputenc}
\usepackage{graphicx}
\usepackage{amsmath}
\usepackage{amsthm}
\usepackage{booktabs,arydshln}
\usepackage{algorithm}
\usepackage{algorithmic}
\usepackage{amssymb}
\usepackage[export]{adjustbox}
\usepackage{multirow}
\usepackage{threeparttable}
\usepackage{tabularx}
\usepackage{color}
\usepackage{tcolorbox}
\usepackage[inline]{enumitem}
\usepackage{tikz}
\usepackage{xparse}
\usepackage{subcaption}
\usetikzlibrary{calc}

\definecolor{red}{rgb}{1.0,0.0,0.0}
\definecolor{orange}{rgb}{1.0,0.65,0.0}
\definecolor{green}{rgb}{0.0,0.75,0.0}
\definecolor{blue}{rgb}{0.0,0.0,1.0}
\definecolor{purple}{rgb}{0.5,0.0,0.5}
\definecolor{white}{rgb}{1.0,1.0,1.0}
\definecolor{redo}{rgb}{1.0,0.0,0.0}
\definecolor{orangeo}{rgb}{1.0,0.65,0.0}
\definecolor{greeno}{rgb}{0.0,1.0,0.0}
\definecolor{blueo}{rgb}{0.0,0.0,1.0}
\definecolor{purpleo}{rgb}{0.5,0.0,0.5}
\definecolor{white}{rgb}{1.0,1.0,1.0}

\newlength\LineWidth
\definecolor{HLcolor}{RGB}{124,18,18}
\sethlcolor{HLcolor}

\newcommand{\mystar}{{\fontfamily{lmr}\selectfont$\star$}}

\makeatletter
\newcommand{\defhighlighter}[3][]{%
  \tikzset{every highlighter/.style={draw=#2, fill opacity=#3, #1}}%
}

\defhighlighter[fill=olive!15]{HLcolor}{.25}

\newcommand{\highlight@DoHighlight}{
  \fill [outer sep = -15pt, inner sep = 0pt, every highlighter, this highlighter,draw=none]
        ($(begin highlight)+(0,8pt)$) rectangle ($(end highlight)+(0,-2pt)$) ;
  \draw[HLcolor,line width=\LineWidth]  ($(begin highlight)+(0,-2pt)$) -- ($(end highlight)+(0,-2pt)$) ;
  \draw[HLcolor,line width=\LineWidth]  ($(begin highlight)+(0,8pt)$) -- ($(end highlight)+(0,8pt)$) ;
}

\newcommand{\highlight@BeginHighlight}{
  \coordinate (begin highlight) at (0,0) ;
}

\newcommand{\highlight@EndHighlight}{
  \coordinate (end highlight) at (0,0) ;
}

\newdimen\highlight@previous
\newdimen\highlight@current

\DeclareRobustCommand*\highlight[1][]{%
  \tikzset{this highlighter/.style={#1}}%
  \SOUL@setup
  \def\SOUL@preamble{%
    \begin{tikzpicture}[overlay, remember picture]
      \highlight@BeginHighlight
      \draw[HLcolor,line width=\LineWidth]  ($(begin highlight)+(0,-2pt)+(0,-0.5\pgflinewidth)$) -- ($(begin highlight)+(0,8pt)+(0,0.5\pgflinewidth)$) ;
      \highlight@EndHighlight
    \end{tikzpicture}%
  }%
  \def\SOUL@postamble{%
    \begin{tikzpicture}[overlay, remember picture]
      \highlight@EndHighlight
      \highlight@DoHighlight
      \draw[HLcolor,line width=\LineWidth]  ($(end highlight)+(0,-2pt)+(0,-0.5\pgflinewidth)$) -- ($(end highlight)+(0,8pt)+(0,0.5\pgflinewidth)$) ;
    \end{tikzpicture}%
  }%
  \def\SOUL@everyhyphen{%
    \discretionary{%
      \SOUL@setkern\SOUL@hyphkern
      \SOUL@sethyphenchar
      \tikz[overlay, remember picture] \highlight@EndHighlight ;%
    }{%
    }{%
      \SOUL@setkern\SOUL@charkern
    }%
  }%
  \def\SOUL@everyexhyphen##1{%
    \SOUL@setkern\SOUL@hyphkern
    \hbox{##1}%
    \discretionary{%
      \tikz[overlay, remember picture] \highlight@EndHighlight ;%
    }{%
    }{%
      \SOUL@setkern\SOUL@charkern
    }%
  }%
  \def\SOUL@everysyllable{%
    \begin{tikzpicture}[overlay, remember picture]
      \path let \p0 = (begin highlight), \p1 = (0,0) in \pgfextra
        \global\highlight@previous=\y0
        \global\highlight@current =\y1
      \endpgfextra (0,0) ;
      \ifdim\highlight@current < \highlight@previous
        \highlight@DoHighlight
        \highlight@BeginHighlight
      \fi
    \end{tikzpicture}%
    \the\SOUL@syllable
    \tikz[overlay, remember picture] \highlight@EndHighlight ;%
  }%
  \SOUL@
}

\def\adl@drawiv#1#2#3{%
        \hskip.5\tabcolsep
        \xleaders#3{#2.5\@tempdimb #1{1}#2.5\@tempdimb}%
                #2\z@ plus1fil minus1fil\relax
        \hskip.5\tabcolsep}
\newcommand{\cdashlinelr}[1]{%
  \noalign{\vskip\aboverulesep
           \global\let\@dashdrawstore\adl@draw
           \global\let\adl@draw\adl@drawiv}
  \cdashline{#1}
  \noalign{\global\let\adl@draw\@dashdrawstore
           \vskip\belowrulesep}}

\DeclareDocumentCommand\MyDBox{O{HLcolor!15}O{HLcolor}m}{%
  \colorlet{HLcolor}{#1!75}
  \highlight[#1!75]{#3}%
}

\title{Multi-Dimensional Explanation of Target Variables from Documents}
\author{
    Diego Antognini,\textsuperscript{\rm 1}
    Claudiu Musat,\textsuperscript{\rm 2}
    Boi Faltings \textsuperscript{\rm 1}
    \\
}
\affiliations{
    \textsuperscript{\rm 1}Ecole Polytechnique Fédérale de Lausanne, Switzerland\\
    \textsuperscript{\rm 2}Swisscom, Switzerland\\
    firstname.lastname@\{epfl.ch, swisscom.com\}\\
}

\begin{document}
\maketitle

\begin{abstract}

Automated predictions require explanations to be interpretable by humans.
Past work used attention and rationale mechanisms to find words that predict the target variable of~a document. Often though, they result in a tradeoff between noisy explanations or a drop in accuracy. 
Furthermore, rationale methods cannot capture the multi-faceted nature of justifications for multiple targets, because of the non-probabilistic nature of the mask.
In this paper, we propose the Multi-Target Masker (MTM) to address these shortcomings.
The novelty lies in the soft multi-dimensional mask that models a relevance probability distribution over the set of target variables to handle ambiguities. 
Additionally, two regularizers guide MTM to induce long, meaningful explanations.
We evaluate MTM on two datasets and show, using standard metrics and human annotations, that the resulting masks are more accurate and coherent than those generated by the state-of-the-art methods. 
Moreover, MTM is the first to also achieve the highest F1 scores for all the target variables simultaneously.

\end{abstract}

\section{Introduction}

\begin{figure}[t]
\centering
\begin{tabular}{@{}c@{\hspace{1mm}}c@{}}
     \underline{Attention Model} & \underline{Multi-Target Masker (Ours)} \\
    Trained on $\ell_{pred}$ & Trained on $\ell_{pred}$ \\
    and no constraint & with $\lambda_p$, $\ell_{sel}$, and $\ell_{cont}$ \\
     \includegraphics[width=0.23\textwidth,height=4.76cm]{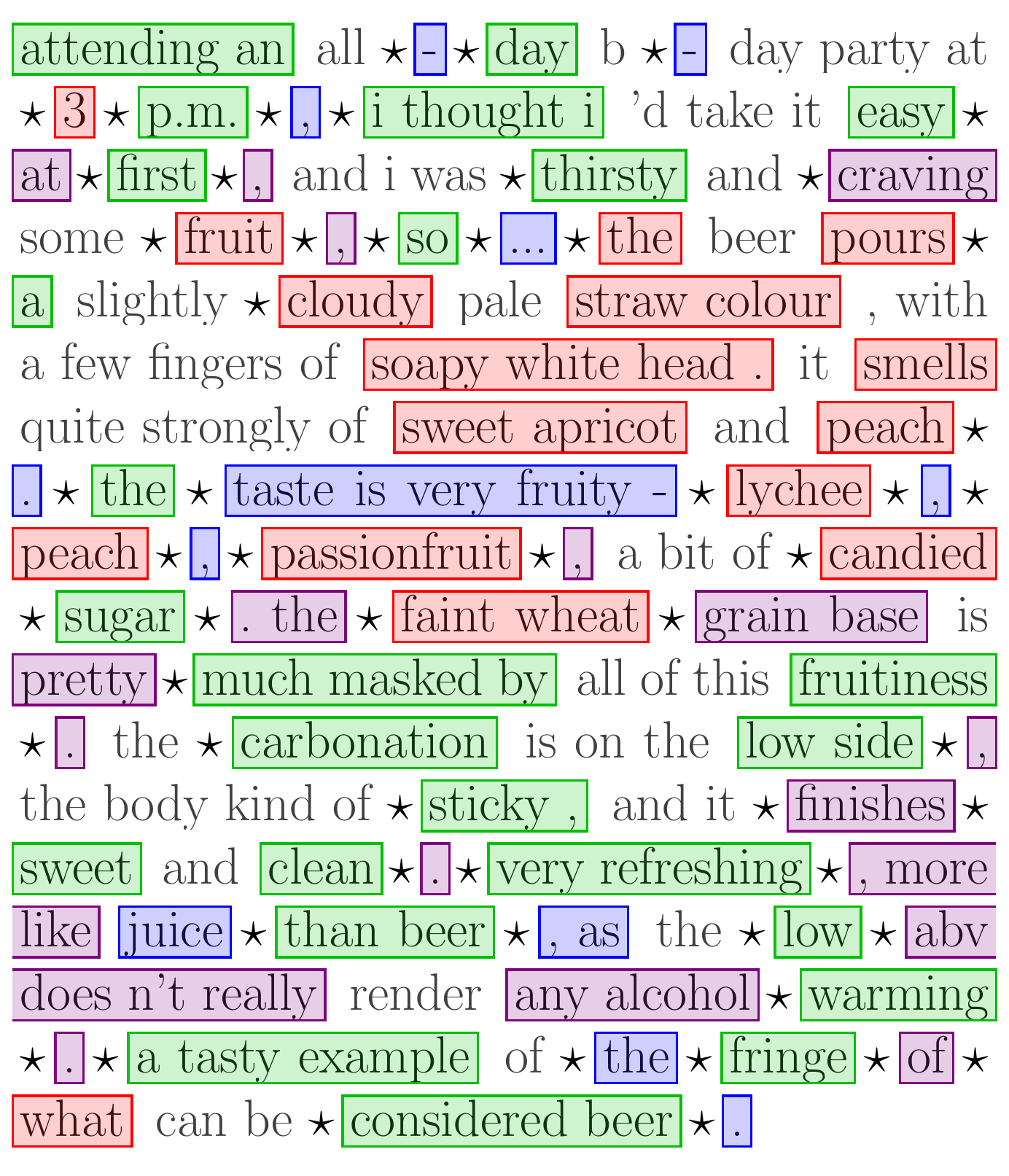} &
     \includegraphics[width=0.23\textwidth,height=4.76cm]{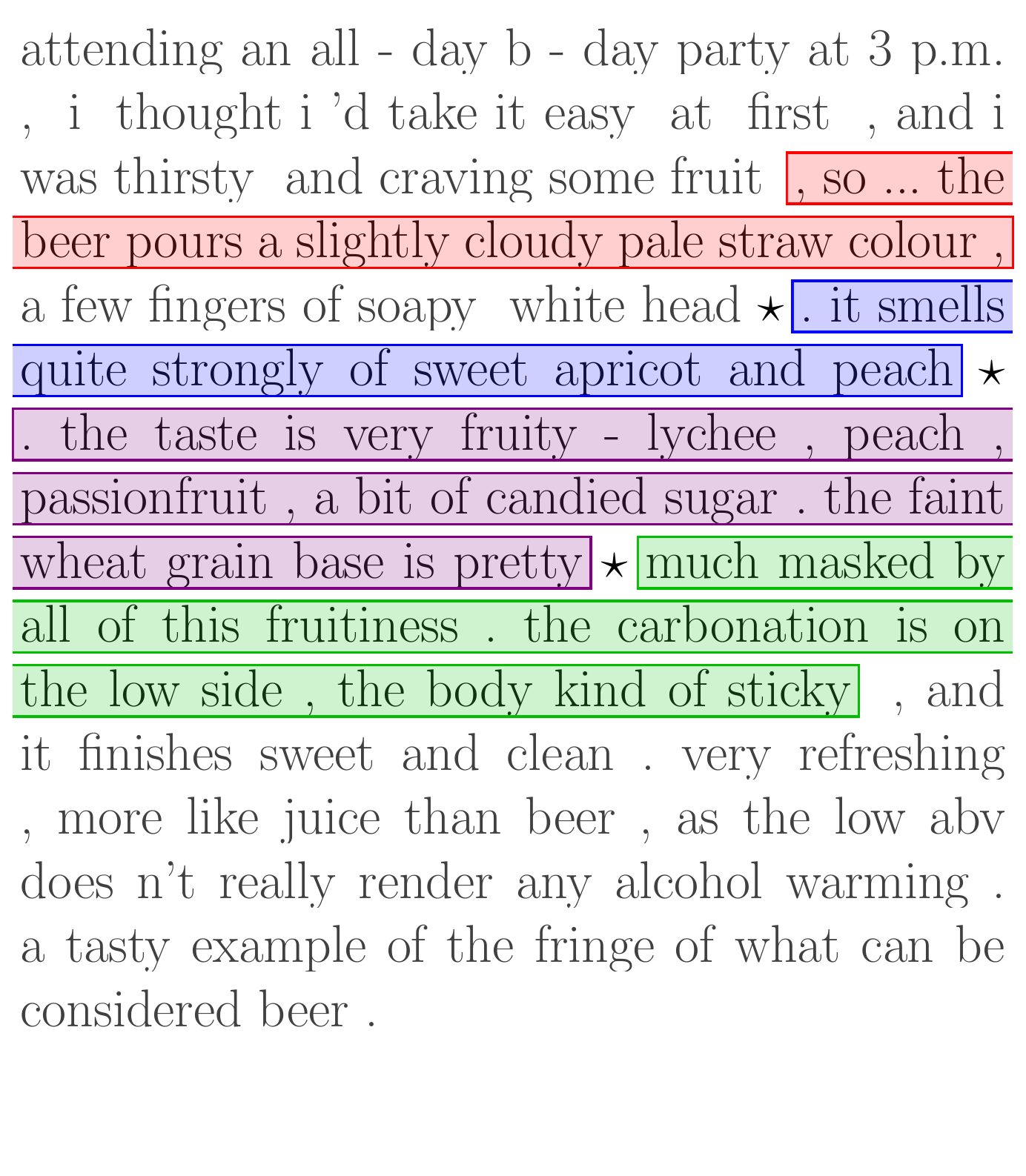}\\
     Aspect Changes \mystar : $56$ & Aspect Changes \mystar : $3$
     \end{tabular}

\caption{\label{sample_lambda}A beer review with explanations produced by an attention model and our Multi-Target Masker model. The colors depict produced rationales (i.e., justifications) of the rated aspects: \MyDBox[red]{\strut Appearance}, \MyDBox[blue]{\strut Smell}, \MyDBox[purple]{\strut Taste}, and \MyDBox[green]{\strut Palate}. 
	The induced rationales mostly lead to long sequences that clearly describe each aspect (one switch \mystar\ per aspect), while the attention model has many short, noisy interleaving sequences.
}
\end{figure}

Neural models have become the standard for natural language processing tasks. Despite the large performance gains achieved by these complex models, they offer little transparency about their inner workings. Thus, their performance comes at the cost of interpretability, limiting their practical utility. Integrating interpretability into a model would supply reasoning for the prediction, increasing its utility.

Perhaps the simplest means of explaining predictions of complex models is by selecting relevant input features. Prior work includes various methods to find relevant words in the text input to predict the target variable of a document. Attention mechanisms \cite{iclr2015,luong-etal-2015-effective} model the word selection by a conditional importance distribution over the inputs, used as explanations to produce a weighted context vector for downstream modules. However, their reliability has been questioned \cite{jain2019attention,pruthi-etal-2020-learning}.
Another line of research includes rationale generation methods \cite{NIPS2017_7062,li2016understanding,lei-etal-2016-rationalizing}. 
If the selected text input features are short and concise -- called a rationale or mask -- and suffice on their own to yield the prediction, it can potentially be understood and verified against domain knowledge \cite{lei-etal-2016-rationalizing,chang2019game}. Specifically, these rationale generation methods have been recently proposed to provide such explanations alongside the prediction.
Ideally, a good rationale should yield the same or higher performance as using the full input.

The key motivation of our work arises from the limitations of the existing methods. First, the attention mechanisms induce an importance distribution over the inputs, but the resulting explanation consists of many short and noisy word sequences (Figure~\ref{sample_lambda}). In addition, the rationale generation methods produce coherent explanations, but the rationales are based on a binary selection of words, leading to the following shortcomings: 
\begin{enumerate*}
	\item they explain only one target variable,
	\item they make a priori assumptions about the data, and 
	\item they make it difficult to capture ambiguities in the text.
\end{enumerate*}
Regarding the first shortcoming, rationales can be multi-faceted by definition and involve support for different outcomes. If that is the case, one has to train, tune, and maintain one model per target variable, which is impractical. For the second, current models are prone to pick up spurious correlations between the input features and the output. Therefore, one has to ensure that the data have low correlations among the target variables, although this may not reflect the real distribution of the data. Finally, regarding the last shortcoming, a strict assignment of words as rationales might lead to ambiguities that are difficult to capture. For example, in an hotel review that states \textit{``The room was large, clean, and close to the beach."},  the word \textit{``room"} refers to the aspects \textit{Room}, \textit{Cleanliness}, and \textit{Location}. All these limitations are implicitly related due to the non-probabilistic nature of the~mask. For further illustrations, see Figure~\ref{sample_new_hotel} and the appendices.

In this work, we take the best of the attention and rationale methods and propose the Multi-Target Masker to address their limitations by replacing the hard binary mask with a soft multi-dimensional mask (one for each target), in an unsupervised and multi-task learning manner, while jointly predicting all the target variables. We are the first to use a probabilistic multi-dimensional mask to explain multiple target variables jointly without any assumptions on the data, unlike previous rationale generation methods. More specifically, for each word, we model a relevance probability distribution over the set of target variables plus the irrelevant case, because many words can be discarded for every target. Finally, we can control the level of interpretability by two regularizers that guide the model in producing long, meaningful rationales.
Compared to existing attention mechanisms, we derive a target importance distribution for each word instead of one over the entire sequence length.

Traditionally, interpretability came at the cost of reduced performance. In contrast, our evaluation shows that on two datasets, in beer and hotel review domains, with up to five correlated targets, our model outperforms strong attention and rationale baselines approaches and generates masks that are strong feature predictors and have a meaningful interpretation. We show that it can be a benefit to:
\begin{enumerate*}
	\item guide the model to focus on different parts of the input text, 
	\item capture ambiguities of words belonging to multiple aspects, and 
	\item further improve the sentiment prediction for all the aspects.
\end{enumerate*}
Thus, interpretability does not come at a cost in our paradigm. 

\section{Related Work}
\subsection{Interpretability}
Developing interpretable models is of considerable interest to the broader research community; this is even more pronounced with neural models \cite{kim2015mind,doshi2017towards}. 
There has been much work with a multitude of approaches in the areas of analyzing and visualizing state activation \cite{KarpathyJL15,li2016visualizing,montavon2018methods}, attention weights \cite{jain2019attention,serrano-smith-2019-attention,pruthi-etal-2020-learning}, and learned sparse and interpretable word vectors \cite{faruqui2015sparse,faruqui-etal-2015-retrofitting,herbelot2015building}.
Other works interpret black box models by locally fitting interpretable models \cite{ribeiro2016should,NIPS2017_7062}.
\cite{li2016understanding} proposed erasing various parts of the input text using reinforcement learning to interpret the decisions. 
However, this line of research aims at providing post-hoc explanations of an already-trained model.
Our work differs from these approaches in terms of what is meant by an explanation and its computation. We defined an explanation as one or multiple text snippets that -- as a substitute of the input text -- are sufficient for the predictions.

\subsection{Attention-based Models}
Attention models \cite{vaswani2017attention,yang2016hierarchical,LinFSYXZB17} have been shown to improve prediction accuracy, visualization, and interpretability. The most popular and widely used attention mechanism is soft attention~\cite{iclr2015}, rather than hard attention~\cite{luong-etal-2015-effective} or sparse ones~\cite{martins2016softmax}.
According to various studies \cite{jain2019attention,serrano-smith-2019-attention,pruthi-etal-2020-learning}, standard attention modules noisily predict input importance; the weights cannot provide safe and meaningful explanations.
Moreover, \cite{pruthi-etal-2020-learning} showed that standard attention modules can fool people into thinking that predictions from a model biased against gender minorities do not rely on the gender.
Our approach differs in two ways from attention mechanisms. First, the loss includes two regularizers to favor long word sequences for interpretability. Second, the normalization is not done over the sequence length but over the target set for each word; each has a relevance probability distribution over the set of target variables.

\subsection{Rationale Models}
The idea of including human rationales during training has been explored in \cite{zhang-etal-2016-rationale,bao-etal-2018-deriving,deyoung-etal-2020-eraser}. Although they have been shown to be beneficial, they are costly to collect and might vary across annotators. In our work, no annotation is needed.

One of the first rationale generation methods was introduced by \cite{lei-etal-2016-rationalizing} in which a generator masks the input text fed to the classifier.
This framework is a cooperative game that selects rationales to accurately predict the label by maximizing the mutual information \cite{chen2018learning}.
\cite{yu-etal-2019-rethinking} proposed conditioning the generator based on the predicted label from a classifier reading the whole input, although it slightly underperformed when compared to the original model \cite{chang2020invariant}.
\cite{chang2019game} presented a variant that generated rationales to perform counterfactual reasoning. Finally, \cite{chang2020invariant} proposed a generator that can decrease spurious correlations in which the selective rationale consists of an extracted chunk of a pre-specified length, an easier variant than the original one that generated the rationale.
In all, these models are trained to generate a hard binary mask as a rationale to explain the prediction of a target variable, and the method requires as many models to train as variables to explain. Moreover, they rely on the assumption that the data have low internal correlations.

In contrast, our model addresses these drawbacks by jointly predicting the rationales of all the target variables (even in the case of highly correlated data) by generating a soft multi-dimensional mask. The probabilistic nature of the masks can handle ambiguities in the induced rationales. In our recent work \cite{antognini2020interacting}, we show~how~to use the induced rationales to generate personalized explanations for recommendation and how human users significantly prefer these over those produced by state-of-the-art models.

\section{The Multi-Target Masker (MTM)}

\begin{figure}[!t]
\centering
\includegraphics[width=.47\textwidth]{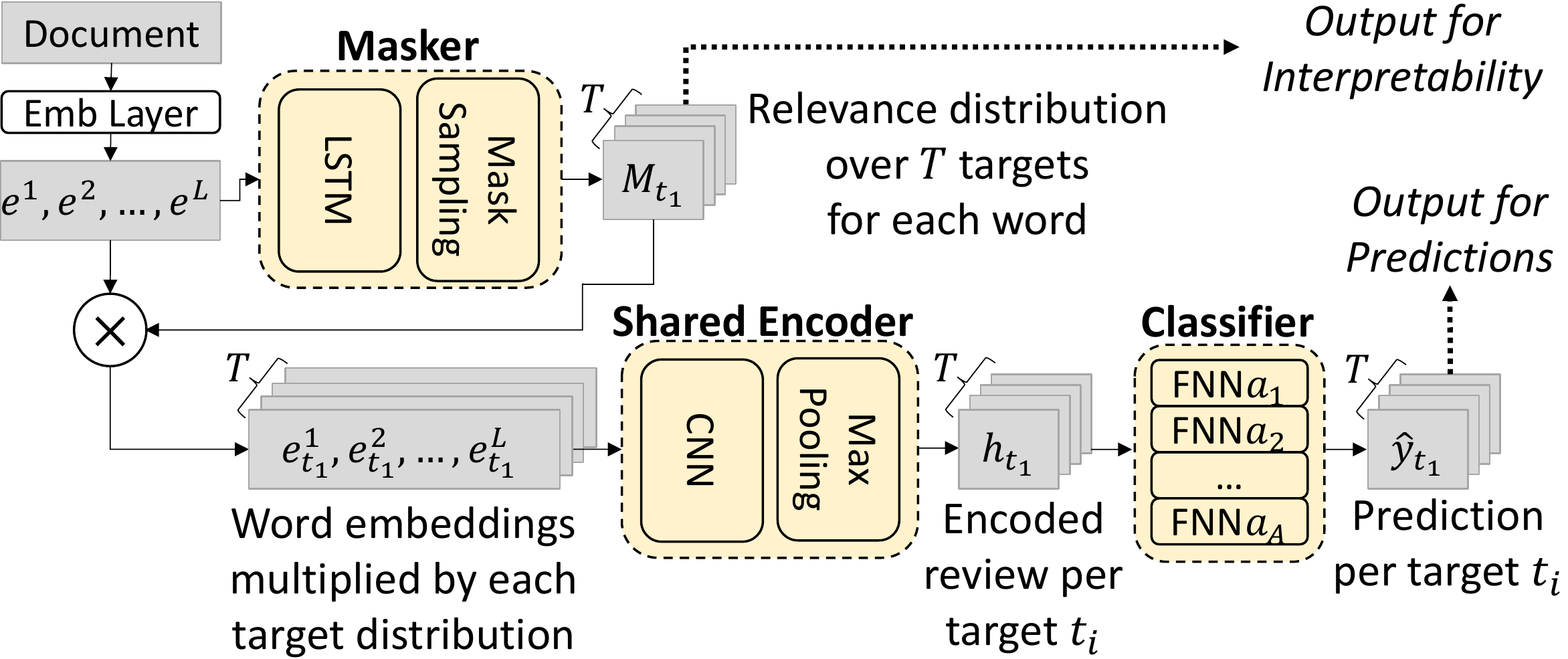}
\caption{\label{model_architecture}The proposed Multi-Target Masker (\textit{MTM}) model architecture to predict and explain $T$ target variables.}
\end{figure}

Let $X$ be a random variable representing a document composed of $L$ words ($x^1, x^2, ..., x^L$), and $Y$ the target $T$-dimensional vector.\footnote{Our method is easily adapted for regression problems.} Our proposed model, called the Multi-Target Masker (MTM), is composed of three components:~1)~a~\textbf{masker} module that computes a probability distribution over the target set for each word, resulting in $T+1$ masks (including one for the irrelevant case); 2)~an \textbf{encoder} that learns a representation of a document $X$ conditioned on the induced masks; 3)~a \textbf{classifier} that predicts the target variables. The overall model architecture is shown in Figure~\ref{model_architecture}. Each module is interchangeable with other models.
\subsection{Model Overview}

\subsubsection{Masker.} The masker first computes a hidden representation $h^\ell$ for each word $x^\ell$ in the input sequence, using their word embeddings $e^1, e^2, ..., e^L$. Many sequence models could realize~this task, such as recurrent, attention, or convolution networks. In our case, we chose a recurrent model to learn the dependencies between the words.
Let $t_i$ be the $i^{th}$ target for $i=1, ..., T$, and~$t_0$ the irrelevant case, because many words are irrelevant to every target. We define the multi-dimensional mask $\mathbf{M} \in \mathbb{R}^{(T+1) \times L}$ as the target relevance distribution $M^\ell \in \mathbb{R}^{(T+1)}$ of each word~$x^\ell$ as follows:
\begin{equation}
    P(\mathbf{M}|X) = \prod_{\ell=1}^L P(M^\ell|x^\ell) = \prod_{\ell=1}^L \prod_{i=0}^T P(m_{t_i}^\ell|x^\ell)
\end{equation}
Because we have categorical distributions, we cannot directly sample $P(M^\ell|x^\ell)$ and backpropagate the gradient through this discrete generation process. Instead, we model the variable $m^\ell_{t_i}$ using the straight through gumbel-softmax~\cite{JangGP17,MaddisonMT17} to approximate sampling from a categorical distribution.\footnote{We also experimented with the implicit reparameterization trick using a Dirichlet distribution \cite{figurnov2018implicit} instead, but we did not obtain a significant improvement.} We model the parameters of each Gumbel-Softmax distribution~$M^\ell$ with a single-layer feed-forward neural network followed by applying a log softmax, which induces the log-probabilities of the $\ell^{th}$  distribution: $\omega_\ell = \text{log}(\text{softmax}(W h^\ell + b))$. $W$ and $b$ are shared across all tokens so that the number of parameters stays constant with respect to the sequence length. We control the sharpness of the distributions with the~temperature~parameter~$\tau$, which dictates the peakiness of the relevance distributions. In our case, we keep the temperature low to enforce the assumption that each word is relevant about one or two targets. Note that compared to attention mechanisms, the word importance is a~probability distribution over~the~targets $\sum_{i=0}^T P(m_{t_i}^\ell|x^\ell) = 1$ instead of a normalization over the sequence length $\sum_{\ell=1}^L P(t^\ell | x^\ell) = 1$.

Given a soft multi-dimensional mask $\mathbf{M} \in \mathbb{R}^{(T+1) \times L}$, we define each sub-mask $M_{t_i} \in \mathbb{R}^L$ as follows:
\begin{equation}
    M_{t_i} = P(m_{t_i}^1|x^1) , P(m_{t_i}^2|x^2) , ... , P(m_{t_i}^L|x^L)
\end{equation}
To integrate the word importance of the induced sub-masks $M_{t_i}$ within the model, we weight the word embeddings by their importance towards a target variable $t_i$, such that $E_{t_i}$ = $E \odot M_{t_i} =$ $e_1 \cdot P(m^1_{t_i}|x^1) , e_2 \cdot P(m^2_{t_i}|x^2), ... , e_L \cdot P(m^L_{t_i}|x^L)$. Thereafter, each modified embedding $E_{t_i}$ is fed into the encoder block. Note that $E_{t_0}$ is ignored because $M_{t_0}$ only serves to absorb probabilities of words that are insignificant.\footnote{if $P(m_{t_0}^\ell|x^\ell) \approx 1.0$, it implies $\sum_{i=1}^T P(m_{t_i}^\ell|x^\ell) \approx 0$ and consequently, $e_{t_i}^\ell \approx \vec{0}$ for  $i=0, ..., T$.}

\subsubsection{Encoder and Classifier.} The encoder includes a convolutional network, followed by max-over-time pooling to obtain a fixed-length feature vector. We chose a convolutional network because it led to a smaller model, faster training, and performed empirically similarly to recurrent and attention models. It produces the fixed-size hidden representation $h_{t_i}$ for each target $t_i$. To exploit commonalities and differences among the targets, we share the weights of the encoder for all $E_{t_i}$. Finally, the classifier block contains for each target variable $t_i$ a two-layer feedforward neural network, followed by a softmax layer to predict the outcome~$\hat{y}_{t_i}$.

\subsubsection{Extracting Rationales.}
To explain the prediction $\hat{y}_{t_i}$ of one target $Y_{t_i}$, we generate its rationale by selecting each word $x^\ell$, whose relevance towards $t_i$ is the most likely: $P(m_{t_i}^\ell|x^\ell) = \max_{j=0, ..., T}{P(m_{t_j}^\ell|x^\ell)}$. Then, we can interpret $P(m_{t_i}^\ell|x^\ell)$ as the model confidence of $x^\ell$ relevant to~$Y_{t_i}$.

\subsection{Enabling the Interpretability of Masks}

The first objective to optimize is the prediction loss, represented as the cross-entropy between the true target label $y_{t_i}$ and the prediction $\hat{y}_{t_i}$ as follows:~\begin{equation}
    \ell_{pred} = \sum_{i=1}^{T} \ell_{cross\_entropy}(y_{t_i}, \hat{y}_{t_i})
\end{equation}
However, training MTM to optimize $\ell_{pred}$ will lead to meaningless sub-masks $M_{t_i}$ because the~model tends to focus on certain words. Consequently, we guide the model to produce long, meaningful word sequences, as shown in Figure~\ref{sample_lambda}.
We propose two regularizers to control the number of selected words and encourage consecutive words to be relevant to the same target. For the first term, we calculate the~probability $p_{sel}$ of tagging a word as relevant to any target as follows:~\begin{equation}
        p_{sel} = \frac{1}{L} \sum_{\ell=1}^L \big( 1 - P(m_{t_0}^\ell|x^\ell) \big)
\end{equation}
We then compute the cross-entropy with a prior hyperparameter $\lambda_p$ to control the expected number of selected words among all target variables, which corresponds to the expectation of a binomial distribution $(p_{sel})$. We minimize the difference between $p_{sel}$ and~$\lambda_p$ as follows:~\begin{equation}
        \ell_{sel} = \ell_{binary\_cross\_entropy}(p_{sel}, \lambda_p)
\end{equation}
The second regularizer discourages the target transition of two consecutive words by minimizing the mean variation of their target distributions, $M^\ell$ and $M^{\ell-1}$. We generalize the formulation of a hard binary selection as suggested by \cite{lei-etal-2016-rationalizing} to a soft probabilistic multi-target selection as follows:\footnote{Early experiments with other distance functions, such as the Kullback–Leibler divergence, produced inferior results.}~\begin{equation}
    \begin{split}
        p_{dis} &= \frac{1}{L} \sum_{\ell=1}^L \frac{\big|\big| M^\ell - M^{\ell-1} \big|\big|_1 }{A + 1}\\
        \ell_{cont} &= \ell_{binary\_cross\_entropy}(p_{dis}, 0)
    \end{split}
\end{equation}
We train our Multi-Target Masker end to end and optimize the loss $\ell_{MTM} = \ell_{pred} + \lambda_{sel} \cdot \ell_{sel} + \lambda_{cont} \cdot \ell_{cont}$, where $\lambda_{sel}$ and $\lambda_{cont}$ control the impact of each~constraint.

\section{Experiments}

We assess our model in two dimensions: the quality of the explanations, obtained from the masks, and the predictive performance. Following previous work \cite{lei-etal-2016-rationalizing,chang2020invariant}, we use sentiment analysis as a demonstration use case, but we extend it to the multi-aspect case. However, we are interested in learning rationales for every aspect at the same time without any prior assumption on the data, where aspect ratings can be highly~correlated.
We first measure the quality of the induced rationales using human aspect sentence-level annotations and an automatic topic model evaluation method. In the second set of experiments, we evaluate MTM on the multi-aspect sentiment classification task in two different domains.\footnote{Code \& data available at \url{https://github.com/Diego999/MTM}.}
\subsection{Experimental Details}
\label{experimental_details}

The review encoder was either a bi-directional recurrent neural network using LSTM~\cite{hochreiter1997long} with $50$ hidden units or a multi-channel text convolutional neural network, similar to~\cite{kim2015mind}, with $3$-, $5$-, and $7$-width filters and $50$ feature maps per filter. Each aspect classifier is a two-layer feedforward neural network with a rectified linear unit activation function~\cite{nair2010rectified}. We used the $200$-dimensional pre-trained word embeddings of~\cite{lei-etal-2016-rationalizing} for beer reviews. For the hotel domain, we trained word2vec~\cite{mikolov2013distributed} on a large collection of hotel reviews~\cite{antognini_hotel_rec} with an embedding size~of~$300$.
We used a dropout~\cite{srivastava2014dropout} of~$0.1$, clipped the gradient norm at $1.0$, added a L2-norm regularizer with a factor of~$10^{-6}$, and trained using early stopping. 
We used~Adam \cite{KingmaB14} with a learning rate of $0.001$. The~temperature $\tau$ for the Gumbel-Softmax distributions was fixed at $0.8$. The two regularizers and the prior of our model were~$\lambda_{sel} = 0.03$, $\lambda_{cont} = 0.03$, and $\lambda_p = 0.15$ for the \textit{Beer} dataset and $\lambda_{sel} = 0.02$, $\lambda_{cont} = 0.02$, and $\lambda_p = 0.10$ for the \textit{Hotel}~one. We ran all experiments for a maximum of $50$ epochs with a batch-size of $256$. We tuned all models on the dev set with 10 random search trials. 

\subsection{Datasets}
\label{sec_datasets}

\begin{table}[t]
\small
\centering
\begin{tabular}{@{}p{4.15cm}@{}c@{\hspace*{2mm}}c@{}}
\multicolumn{1}{c}{\bf Dataset} & \multicolumn{1}{c}{\bf Beer}  & \multicolumn{1}{c}{\bf Hotel}\\
\toprule
Number of reviews& $1,586,259$ & $140,000$\\
Average words per review& $147.1 \pm 79.7$  & $188.3 \pm 50.0$\\
Average sentences per review& $10.3 \pm 5.4$  & $10.4 \pm 4.4$\\
Number of Aspects & $4$ & $5$ \\
Avg./Max corr. between aspects & $71.8\% / 73.4\%$ & $63.0\% / 86.5\%$\\\end{tabular}
\caption{\label{app_dataset_description_main}Statistics of the multi-aspect review datasets. Both datasets have high correlations between aspects.}
\end{table}

\cite{McAuley2012} provided $1.5$~million English beer reviews from BeerAdvocat. Each contains multiple sentences describing various beer aspects: \textit{Appearance}, \textit{Smell}, \textit{Palate}, and \textit{Taste}; users also provided a five-star rating for each aspect. 
To evaluate the robustness of the models across domains, we sampled $140\,000$ hotel reviews from~\cite{antognini_hotel_rec}, that contains $50$~million reviews from TripAdvisor. Each review contains a five-star rating for each aspect: \textit{Service}, \textit{Cleanliness}, \textit{Value}, \textit{Location}, and \textit{Room}. The descriptive statistics are shown in Table~\ref{app_dataset_description_main}.

There are high correlations among the rating scores of different aspects in the same review ($71.8\%$ and $63.0\%$ on average for the beer and hotel datasets, respectively). This makes it difficult to directly learn textual justifications for single-target rationale generation models \cite{chang2020invariant,chang2019game,lei-etal-2016-rationalizing}. Prior work used~separate decorrelated train sets for each aspect and excluded~aspects with a high correlation, such as \textit{Taste}, \textit{Room}, and \textit{Value}. However, these assumptions do not reflect the real data distribution. Therefore, we keep the original data (and thus can show that our model does not suffer from the high correlations). We binarize the problem as in previous work \cite{bao-etal-2018-deriving,chang2020invariant}: ratings at three and above are labeled as positive and the rest as negative. We split the data into $80/10/10$ for the train, validation, and test sets.
Compared to the beer reviews, the hotel ones were longer, noisier, and less structured, as shown in Appendices~\ref{samples_with_visualization} and \ref{app_decorr}. 

\subsection{Baselines}
\label{sec_baselines}

We compare our Multi-Target Masker~(\textit{MTM}) with various baselines. We group them in three levels of interpretability:
\begin{itemize}
    \item \textit{None}. We cannot extract the input features the model used to make the predictions;
    \item \textit{Coarse-grained}. We can observe what parts of the input a model used to discriminate all aspect sentiments without knowing what part corresponded to what aspect;
    \item \textit{Fine-grained}. For each aspect, a model selects input features to make the prediction.
\end{itemize}

We first use a simple baseline, \textit{SENT}, that reports the majority sentiment across the aspects, as the aspect ratings are highly correlated. Because this information is not available at testing, we trained a model to predict the majority sentiment of a review as suggested by \cite{wang-manning-2012-baselines}. The second baseline we used is a shared encoder followed by $T$~classifiers that we denote \textit{BASE}. These models do not offer any interpretability. We extend it with~a shared attention mechanism~\cite{iclr2015} after the encoder, noted as \textit{SAA} in our study, that provides a coarse-grained interpretability; for all aspects, \textit{SAA} focuses on the same words in the~input.

Our final goal is to achieve the best performance and provide fine-grained interpretability in order to visualize what sequences of words a model focuses on to predict the aspect sentiments.
To this end, we include other baselines: two trained \textit{separately} for each aspect (e.g., current rationale models) and two trained with a \textit{multi-aspect} sentiment loss. 
For the first ones, we employ the the well-known \textit{NB-SVM} \cite{wang-manning-2012-baselines} for sentiment analysis tasks, and we then use the Single-Aspect Masker (\textit{SAM}) \cite{lei-etal-2016-rationalizing}, each trained separately for each aspect.

The two last methods contain a separate encoder, attention mechanism, and classifier~for each aspect. We utilize two types of attention mechanisms, additive~\cite{iclr2015} and sparse~\cite{martins2016softmax}, as sparsity in the attention has been shown to induce useful, interpretable representations. We call them Multi-Aspect Attentions~(\textit{MAA}) and Sparse-Attentions~(\textit{MASA}), respectively. Diagrams of the baselines can be found in Appendix~\ref{baseline_architecture}.

We demonstrate that the induced sub-masks $M_{t_i}, ..., M_{t_T}$ computed from \textit{MTM}, bring fine-grained interpretability and are meaningful for other models to improve performance. To do so, we extract and concatenate the masks to the word embeddings, resulting in contextualized embeddings~\cite{peters-etal-2018-deep}, and train \textit{BASE} with those. We call this variant \textit{MTM\textsuperscript{C}}, that is smaller and has faster inference than~\textit{MTM}.

\section{Results}

\subsection{Multi-Rationale Interpretability}

We first verify whether the inferred rationales of \textit{MTM} are meaningful and interpretable, compared to the other models.

\subsubsection{Precision.}
\label{subsub_prec}
Evaluating explanations that consist of coherent pieces of text is challenging because there is no gold standard for reviews. \cite{McAuley2012} have provided $994$ beer reviews with sentence-level aspect annotations (although our model computes masks at a finer level). Each sentence was annotated with one aspect label, indicating what aspect that sentence covered. We evaluate the precision of the words selected by~each model, as in \cite{lei-etal-2016-rationalizing}.
We use trained models on the \textit{Beer} dataset and extracted a similar number of selected words for a fair comparison. We also report the results of the models from \cite{lei-etal-2016-rationalizing}: \textit{NB-SVM}, the Single-Aspect Attention and Masker (\textit{SAA} and \textit{SAM}, respectively); they use the separate decorrelated train sets for each aspect because they compute hard masks.\footnote{When trained on the original data, they performed significantly worse, showing the limitation in handling correlated variables.}

Table~\ref{perfs_precision_beer} presents the precision of the masks and attentions computed on the sentence-level aspect annotations. We show that the generated sub-masks obtained with our Multi-Target Masker (\textit{MTM}) correlates best with the human judgment. In comparison to \textit{SAM}, the \textit{MTM} model obtains significantly~higher precision with an average of $+1.13$. Interestingly, \textit{NB-SVM} and attention models (\textit{SAA}, \textit{MASA}, and \textit{MAA}) perform  poorly compared with the mask models, especially \textit{MASA}, which focuses only on a couple of words due to the sparseness of the attention. In Appendix~\ref{app:impact_length}, we also analyze the impact of the length of the explanations.

\subsubsection{Semantic Coherence.}
\label{sec_mask_coh}

\begin{table}[t]
\small
\centering
\begin{threeparttable}
    \centering
\begin{tabular}{@{}lc@{\hspace{2mm}}c@{\hspace{1mm}}c@{}}
 & \multicolumn{3}{c}{\textbf{Precision / \% Highlighted Words}}\\
\cmidrule(lr){2-4}
\bf Model & \textit{Smell} & \textit{Palate} & \textit{Appearance}\\
\toprule
NB-SVM\tnote{*} & $21.6$ / $7\%$ & $24.9$ / $7\%$ & $38.3$ / $13\%$\\
SAA\tnote{*} & $88.4$ / $7\%$ & $65.3$ / $7\%$ & $80.6$ / $13\%$\\
SAM\tnote{*} & $95.1$ / $7\%$ & $80.2$ / $7\%$ & $96.3$ / $14\%$\\
MASA & $87.0$ / $4\%$ & $42.8$ / $5\%$ & $74.5$ / $\ \ 4\%$\\
MAA & $51.3$ / $7\%$ & $32.9$ / $7\%$ & $44.9$ / $14\%$\\
\textbf{MTM} & $\mathbf{96.6}$ / $7\%$ & $\mathbf{81.7}$ / $7\%$ & $\mathbf{96.7}$ / $14\%$\\
\end{tabular}
\begin{tablenotes}
     \item[*]\small Model trained separately for each aspect.
   \end{tablenotes}
\end{threeparttable}
\caption{\label{perfs_precision_beer}Performance related to human evaluation, showing the precision of the selected words for each aspect of the \textit{Beer} dataset. The percentage of words indicates the number of highlighted words of the full review.}
\end{table}

In addition to evaluating the rationales with human annotations, we compute their semantic interpretability. According to \cite{aletras2013evaluating,lau2014machine}, normalized point mutual information (NPMI) is a good metric for the qualitative evaluation of topics because it matches human judgment most closely. However, the top-$N$ topic words used for evaluation are often selected arbitrarily. To alleviate this problem, we followed \cite{lau2016sensitivity}. We compute the topic coherence over several cardinalities and report the results and average (see Appendix~\ref{sec_topic_words}); those authors claimed that the mean leads to a more stable and robust evaluation.

\begin{table}[t]
\small
\centering
\begin{threeparttable}[t]
    \centering
\begin{tabular}{
@{}l@{\hspace*{1.75mm}}
c@{\hspace*{1.75mm}}
c@{\hspace*{1.75mm}}
c@{\hspace*{1.75mm}}
c@{\hspace*{1.75mm}}
c@{\hspace*{1.75mm}}
c@{\hspace*{1.75mm}}
c@{}}
& \multicolumn{7}{c}{\textbf{NPMI}}\\
\cmidrule(lr){2-8}
\textbf{Model} & $N=5$ & $10$ & $15$ & $20$ & $25$ & $30$ & \ Mean\tnote{\textsuperscript{\textdagger}}\\
\toprule
\multicolumn{8}{c}{\textit{Beer}} \\
SAM\tnote{*} & $0.046$ & $0.120$ & $0.129$ & $0.243$ & $0.308$ & $0.396$ & \ $0.207$ \\
MASA & $0.020$ & $0.082$ & $0.130$ & $0.168$ & $0.234$ & $0.263$ & \ $0.150$\\
MAA & $0.064$ & $\mathbf{0.189}$ & $0.255$ & $0.273$ & $0.332$ & $0.401$ & \ $0.252$ \\
\textbf{MTM} & $\mathbf{0.083}$ & $0.187$ & $\mathbf{0.264}$ & $\mathbf{0.348}$ & $\mathbf{0.477}$ & $\mathbf{0.410}$ & \ $\mathbf{0.295}$ \\
\midrule
\multicolumn{8}{c}{\textit{Hotel}} \\
SAM\tnote{*} & $0.041$ & $0.103$ & $0.152$ & $0.180$ & $0.233$ & $0.281$ & \ $0.165$ \\
MASA & $0.043$ & $0.127$ & $0.166$ & $0.295$ & $0.323$ & $0.458$ & \ $0.235$\\
MAA & $0.128$ & $0.218$ & $\mathbf{0.352}$ & $0.415$ & $0.494$ & $0.553$ & \ $0.360$\\
\textbf{MTM} & $\mathbf{0.134}$ & $\mathbf{0.251}$ & $0.349$ & $\mathbf{0.496}$ & $\mathbf{0.641}$ & $\mathbf{0.724}$ & \ $\mathbf{0.432}$\\
\end{tabular}
\begin{tablenotes}
     \item[*]\small Model trained separately for each aspect.
     \item[\textsuperscript{\textdagger}]\small The metric that correlates best with human judgment \cite{lau2016sensitivity}.
   \end{tablenotes}
\end{threeparttable}
\caption{\label{topics_perfs}Performance on automatic evaluation, showing the average topic coherence (NPMI) across different top-$N$ words for each dataset. We considered each aspect $a_i$ as a topic and used the masks/attentions to compute~$P(w|a_i)$.}
\begin{tikzpicture}[overlay]

\node [coordinate, shift={(3.2cm, 7.7cm)},inner sep=0pt](c1){};
\node [coordinate,shift={(3.2cm, 3.2cm)},inner sep=0pt](c2){};
\node [coordinate,shift={(4.125cm, 3.2cm)},inner sep=0pt](c3){};
\node [coordinate,shift={(4.125cm, 7.7cm)},inner sep=0pt](c4){};
\draw [fill=gray, fill opacity=0.1, draw=gray,rounded corners = 1ex] (c1) -- (c2) -- (c3) -- (c4) -- cycle; 

\end{tikzpicture}
\end{table}

The results are shown in~Table~\ref{topics_perfs}. We show that the computed masks by \textit{MTM} lead to the highest mean NPMI and, on average, $20\%$  superior results in both datasets, while only needing a single training. Our \textit{MTM} model significantly outperforms \textit{SAM} and the attention models (\textit{MASA} and \textit{MAA}) for $N \ge 20$ and $N=5$. For $N=10$ and $N=15$, \textit{MTM} obtains~higher scores in two out of four cases ($+.033$ and $+.009$). For the other two, the difference was below $.003$. SAM obtains poor results in all cases. 

We analyzed the top words for each aspect by conducting a human evaluation to identify intruder words (i.e., words not matching the corresponding aspect). Generally, our model found better topic words: approximately $1.9$ times fewer intruders than other methods for each aspect and each dataset. More details are available in Appendix~\ref{sec_topic_words}.

\subsection{Multi-Aspect Sentiment Classification}
\label{masc}

\begin{table*}[t]
\small
\centering
\begin{tabular}
{@{}cl>{}ll@{}>{}c@{}>{}c@{}>{}c@{\hspace*{2mm}}>{}c@{\hspace*{2mm}}>{}c@{\hspace*{2mm}}>{}c@{\hspace*{2mm}}>{}c@{\hspace*{2mm}}>{}c@{}}
& & & & & \multicolumn{5}{c}{\textbf{F1 Scores}}\\
\cmidrule(lr){6-10}
\multirow{13}{*}{\rotatebox[origin=c]{90}{\centering \textit{Beer Reviews}}} &  \multicolumn{1}{c}{\bf Interp.} & \multicolumn{2}{c}{\bf Model} & \multicolumn{1}{c}{\bf Params} & \multicolumn{1}{c}{\bf Macro} & \multicolumn{1}{c}{\boldmath$A_1$} & \multicolumn{1}{c}{\boldmath$A_2$} & \multicolumn{1}{c}{\boldmath$A_3$} & \multicolumn{1}{c}{\boldmath$A_4$}\\
\cmidrule[0.08em]{2-10}
& \multicolumn{1}{c}{\multirow{2}{*}{\parbox{1.cm}{\centering None}}}
& SENT & Sentiment Majority & $560k$ & $73.01$ & $71.83$ & $75.65$ & $71.26$ & $73.31$\\
& & BASE & $\text{Emb}_{200}$ + $\text{Enc}_\textsubscript{CNN}$ + \text{Clf} & $188k$ & $76.45$ & $71.44$ & $78.64$ & $74.88$ & $80.83$\\
\cmidrule{3-10}
& \multicolumn{1}{c}{\multirow{2}{*}{\parbox{1.cm}{\centering Coarse-grained}}}
& \multirow{2}{*}{SAA} & $\text{Emb}_{200}$ + $\text{Enc}_\textsubscript{CNN}$ + $\text{A}_\textsubscript{Shared}$ + \text{Clf} & $226k$ & $77.06$ & $73.44$ & $78.68$ & $75.79$ & $80.32$\\
& & & $\text{Emb}_{200}$ + $\text{Enc}_\textsubscript{LSTM}$ + $\text{A}_\textsubscript{Shared}$ + \text{Clf} & $219k$ & $78.03$ & $74.25$ & $79.53$ & $75.76$ & $82.57$\\
\cmidrule{3-10}
& \multicolumn{1}{c}{\multirow{6}{*}{\parbox{1.cm}{\centering Fine-grained}}}
& NB-SVM & \cite{wang-manning-2012-baselines} & $4 \cdot 560k$ & $72.11$ & $72.03$ & $74.95$ & $68.11$ & $73.35$\\
& & SAM & \cite{lei-etal-2016-rationalizing} & $4 \cdot 644k$ & $76.62$ & $72.93$ & $77.94$ & $75.70$ & $79.91$\\
& & MASA & $\text{Emb}_{200}$ + $\text{Enc}_\textsubscript{LSTM}$ + $\text{A}^\textsubscript{Sparse}_\textsubscript{Aspect-wise}$ + \text{Clf} & $611k$ & $77.62$ & $72.75$ & $79.62$ & $75.81$ & $82.28$\\
& & MAA & $\text{Emb}_{200}$ + $\text{Enc}_\textsubscript{LSTM}$ + $\text{A}_\textsubscript{Aspect-wise}$ + \text{Clf} & $611k$ & $78.50$ & $74.58$ & $79.84$ & $77.06$ & $82.53$\\
\cmidrule{3-10}
& & MTM\textsuperscript{} & $\text{Emb}_{200}$ + \text{Masker} + $\text{Enc}_\textsubscript{CNN}$ + \text{Clf} (Ours) & $289k$ & $78.55$ & $74.87$ & $79.93$ & $77.39$ & $82.02$\\
& & \textbf{MTM\textsuperscript{C}} & \textbf{$\text{Emb}_{200+4}$ +  $\text{Enc}_\textsubscript{CNN}$ + \text{Clf} (Ours)} & $191k$ & $\mathbf{78.94}$ & $\mathbf{75.02}$ & $\mathbf{80.17}$ & $\mathbf{77.86}$ & $\mathbf{82.71}$\\
\end{tabular}
\begin{tabular}
{@{}cl>{}ll@{}>{}c@{}>{}c@{}>{}c@{\hspace*{2mm}}>{}c@{\hspace*{2mm}}>{}c@{\hspace*{2mm}}>{}c@{\hspace*{2mm}}>{}c@{}}
 & & & & & \multicolumn{6}{c}{\textbf{F1 Scores}}\\
\cmidrule(lr){6-11}
\multirow{14}{*}{\rotatebox[origin=c]{90}{\centering \textit{Hotel Reviews}}} & \multicolumn{1}{c}{\bf Interp.} & \multicolumn{2}{c}{\bf Model} & \multicolumn{1}{c}{\bf Params} & \multicolumn{1}{c}{\bf Macro} & \multicolumn{1}{c}{\boldmath$A_1$} & \multicolumn{1}{c}{\boldmath$A_2$} & \multicolumn{1}{c}{\boldmath$A_3$} & \multicolumn{1}{c}{\boldmath$A_4$} & \multicolumn{1}{c}{\boldmath$A_5$}\\
\cmidrule[0.08em]{2-11}
& \multicolumn{1}{c}{\multirow{2}{*}{\parbox{1.cm}{\centering None}}}
& SENT & Sentiment Majority& $309k$ & $85.91$ & $89.98$ & $90.70$ & $92.12$ & $65.09$ & $91.67$\\
& & BASE & $\text{Emb}_{300}$ + $\text{Enc}_\textsubscript{CNN}$ + \text{Clf} & $263k$ & $90.30$ & $92.91$ & $93.55$ & $94.12$ & $76.65$ & $94.29$\\
\cmidrule(lr){3-11}
& \multicolumn{1}{c}{\multirow{2}{*}{\parbox{1.cm}{\centering Coarse-grained}}}
& \multirow{2}{*}{SAA} & $\text{Emb}_{300}$ + $\text{Enc}_\textsubscript{CNN}$ + $\text{A}_\textsubscript{Shared}$ + \text{Clf} & $301k$ & $90.12$ & $92.73$ & $93.55$ & $93.76$ & $76.40$ & $94.17$\\
& & & $\text{Emb}_{300}$ + $\text{Enc}_\textsubscript{LSTM}$ + $\text{A}_\textsubscript{Shared}$ + \text{Clf} & $270k$ & $88.22$ & $91.13$ & $92.19$ & $93.33$ & $71.40$ & $93.06$\\
\cmidrule(lr){3-11}
& \multicolumn{1}{c}{\multirow{6}{*}{\parbox{1.cm}{\centering Fine-grained}}}

& NB-SVM & \cite{wang-manning-2012-baselines} & $5 \cdot 309k$ & $87.17$ & $90.04$ & $90.77$ & $92.30$ & $71.27$ & $91.46$\\

& & SAM & \cite{lei-etal-2016-rationalizing} & $5\cdot824k$ & $87.52$ & $91.48$ & $91.45$ & $92.04$ & $70.80$ & $91.85$\\
& & MASA & $\text{Emb}_{200}$ + $\text{Enc}_\textsubscript{LSTM}$ + $\text{A}^\textsubscript{Sparse}_\textsubscript{Aspect-wise}$ + \text{Clf} & $1010k$ & $90.23$ & $93.11$ &    $93.32$ & $93.58$ & $77.21$ & $93.92$ \\
& & MAA & $\text{Emb}_{300}$ + $\text{Enc}_\textsubscript{LSTM}$ + $\text{A}_\textsubscript{Aspect-wise}$ + \text{Clf} & $1010k$ & $90.21$ & $92.84$ & $93.34$ & $93.78$ & $76.87$ & $94.21$\\
\cdashlinelr{3-11}
& & \multirow{1}{*}{MTM} & $\text{Emb}_{300}$ + \text{Masker} + $\text{Enc}_\textsubscript{CNN}$ + \text{Clf} (Ours)& $404k$ & $89.94$ & $92.84$ & $92.95$ & $93.91$ & $76.27$ & $93.71$\\
& & \textbf{MTM\textsuperscript{C}} & \textbf{$\text{Emb}_{300+5}$ + $\text{Enc}_\textsubscript{CNN}$ + \text{Clf} (Ours)} & $267k$ & $\mathbf{90.79}$ & $\mathbf{93.38}$ & $\mathbf{93.82}$ & $\mathbf{94.55}$ & $\mathbf{77.47}$ & $\mathbf{94.71}$ \\
\end{tabular}
\caption{\label{perfs_full_beer}Performance of the multi-aspect sentiment classification task for the \textit{Beer} (top) and \textit{Hotel} (bottom) datasets.}
\end{table*}

We showed that the inferred rationales of \textit{MTM} were significantly more accurate and semantically coherent than those produced by the other models. Now, we inquire as to whether the masks could become a benefit rather than a cost in performance for the multi-aspect sentiment classification.

\subsubsection{Beer Reviews.}
\label{sec_beer_reviews}

We report the macro F1 and individual score for each aspect $A_i$. Table~\ref{perfs_full_beer} (top) presents the results for the \textit{Beer} dataset. The Multi-Target Masker (\textit{MTM}) performs better on average than all the baselines and provided fine-grained interpretability. Moreover, \textit{MTM} has two times fewer parameters than the aspect-wise attention models.

The contextualized variant \textit{MTM\textsuperscript{C}} achieves a macro F1 score absolute improvement of $0.44$~and $2.49$ compared~to \textit{MTM} and \textit{BASE}, respectively. These results highlight that the inferred masks are meaningful to improve the performance while bringing fine-grained interpretability to \textit{BASE}. It is $1.5$ times smaller than \textit{MTM} and has a faster inference.

\textit{NB-SVM}, which offers fine-grained interpretability and was trained separately for each aspect, significantly underperforms when compared to \textit{BASE} and, surprisingly, to \textit{SENT}. As shown in Table~\ref{app_dataset_description_main}, the sentiment correlation between any pair of aspects of the \textit{Beer} dataset is on average $71.8\%$. Therefore, by predicting the sentiment of one~aspect correctly, it is likely that other aspects share the same~polarity. We suspect that the linear model \textit{NB-SVM} cannot~capture the correlated relationships between aspects, unlike the non-linear (neural) models that have a higher capacity. The shared attention models perform better than \textit{BASE} but provide only coarse-grained interpretability. \textit{SAM} is outperformed by all the models except \textit{SENT}, \textit{BASE}, and \textit{NB-SVM}. 

\begin{figure}[!t]
\centering

\begin{tabular}{@{}l@{}}
	\includegraphics[width=0.5\textwidth]{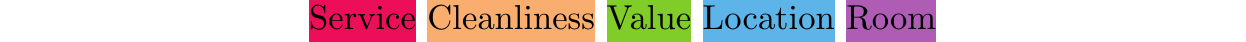} \\
	 \multicolumn{1}{c}{\underline{Multi-Target Masker (Ours)}}\\
     \includegraphics[width=0.4675\textwidth,height=1.88cm]{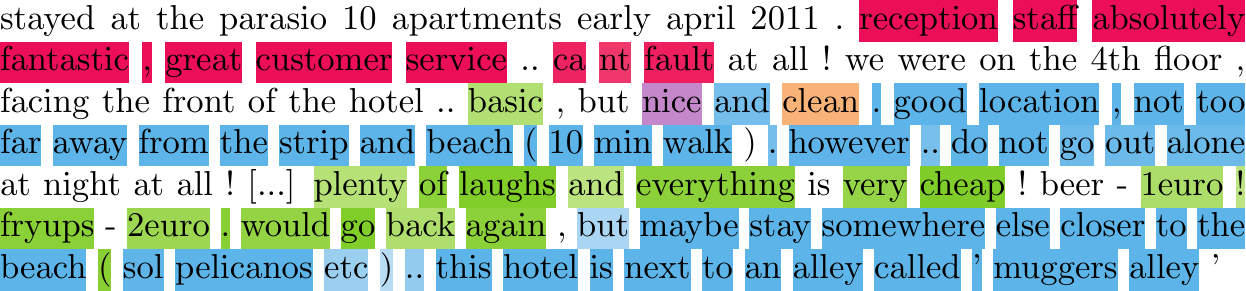} \\
     \multicolumn{1}{c}{\underline{Single-Aspect Masker}}\\
     \includegraphics[width=0.4675\textwidth,height=1.88cm]{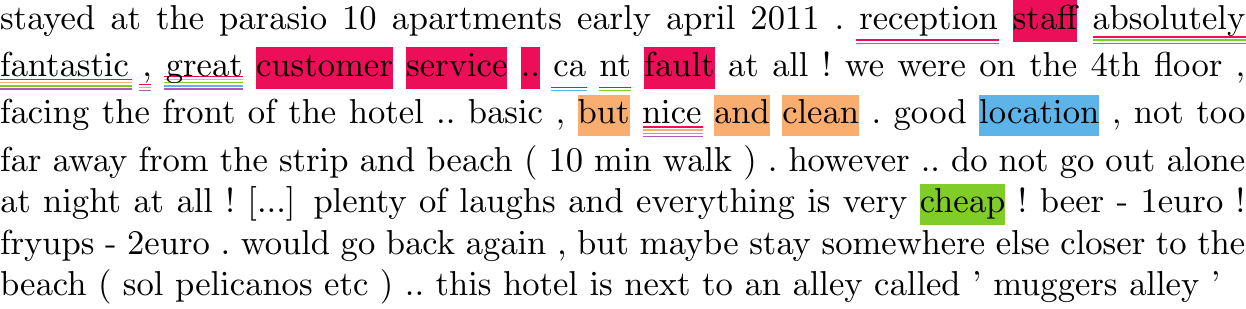} \\
     \multicolumn{1}{c}{\underline{Multi-Aspect Attentions}}\\
     \includegraphics[width=0.4675\textwidth,height=1.88cm]{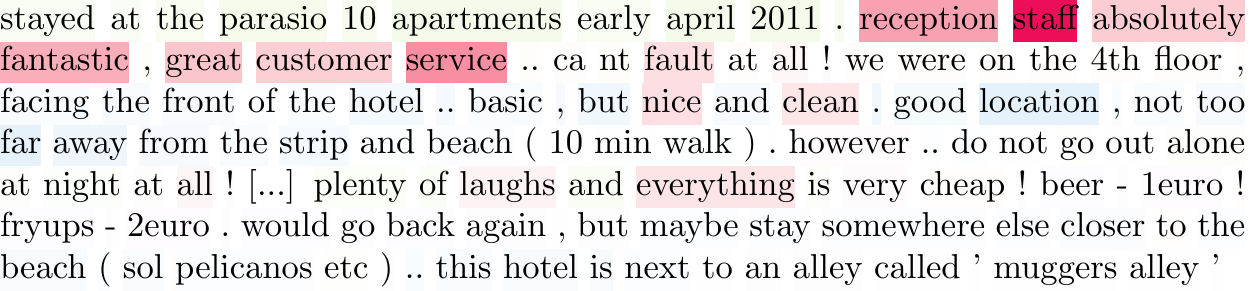} \\
     \multicolumn{1}{c}{\underline{Multi-Aspect Sparse-Attentions}}\\
     \includegraphics[width=0.4675\textwidth,height=1.88cm]{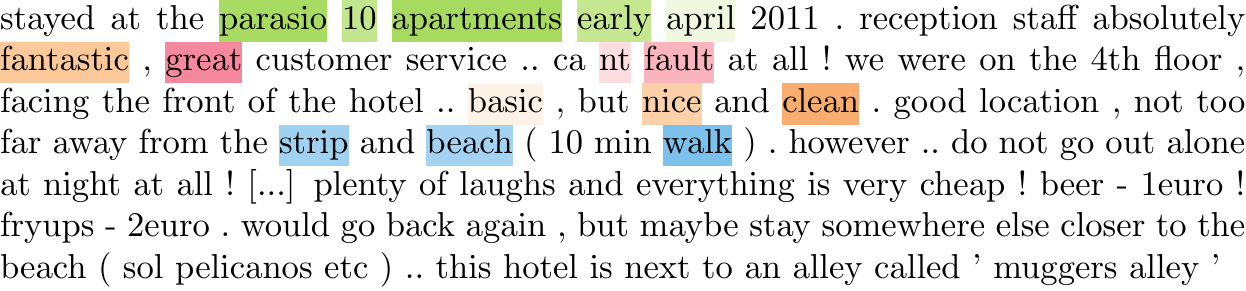}\\
\end{tabular}

\caption{\label{sample_new_hotel}Induced rationales on a truncated hotel review, where shade colors represent the model confidence towards the aspects. \textit{MTM} finds most of the crucial spans of words with a small amount of noise. \textit{SAM} lacks coverage but identifies words where half are correct and the others ambiguous (represented with colored underlines).}
\end{figure}

\subsubsection{Model Robustness - Hotel Reviews.}
We check the robustness of our model on another domain. Table~\ref{perfs_full_beer} (bottom) presents the results of the \textit{Hotel} dataset. The contextualized variant \textit{MTM\textsuperscript{C}}~outperforms all other models significantly with a macro F1 score improvement of $0.49$. Moreover, it achieves the best individual F1 score for each aspect~$A_i$. This shows that the learned mask $\mathbf{M}$ of \textit{MTM} is again meaningful because it increases the performance and adds interpretability to \textit{BASE}.
Regarding \textit{MTM}, we see that it performs slightly worse than the aspect-wise attention models  \textit{MASA} and \textit{MAA} but has $2.5$ times fewer~parameters.

A visualization of a truncated hotel review with the extracted rationales and attentions is available in Figure~\ref{sample_new_hotel}. Not only do probabilistic masks enable higher performance, they better capture parts of reviews related to each aspect compared to other methods. More samples of beer and hotel reviews can be found in Appendix~\ref{full_beer_sample}.

To summarize, we have shown that the regularizers in \textit{MTM} guide the model to produce high-quality masks as explanations while performing slightly better than the strong attention models in terms of prediction performance. However, we demonstrated that including the inferred masks into word embeddings and training a simpler model achieved the best performance across two datasets and and at the same time, brought fine-grained interpretability. Finally, \textit{MTM} supported high correlation among multiple target variables.

\subsubsection{Hard Mask versus Soft Masks.}
\label{sec_decorr}

\textit{SAM} is the neural model that obtained the lowest relative macro F1 score in the two datasets compared with \textit{MTM\textsuperscript{C}}: a difference of $-2.32$ and $-3.27$ for the \textit{Beer} and \textit{Hotel} datasets, respectively. Both datasets have a high average correlation between the aspect ratings: $71.8\%$ and $63.0\%$, respectively (see Table~\ref{app_dataset_description_main}).
Therefore, it makes it challenging for rationale models to learn the justifications of the aspect ratings directly. Following the observations of \cite{lei-etal-2016-rationalizing,chang2019game,chang2020invariant}, this highlights that single-target rationale models suffer from high correlations and require data to satisfy certain constraints, such as low correlations. In contrast, \textit{MTM} does not require any particular assumption on the data.

We compare \textit{MTM} in a setting where the aspect ratings were less correlated, although it does not reflect the real distribution of the aspect ratings. We employ the decorrelated subsets of the \textit{Beer} reviews from \cite{lei-etal-2016-rationalizing,chang2020invariant}. It has an average correlation of $27.2\%$ and the aspect \textit{Taste} is removed. 

We find similar trends but stronger results: \textit{MTM} significantly generates better rationales and achieves higher F1 scores than \textit{SAM} and the attention models. The contextualized variant \textit{MTM\textsuperscript{C}} further improves the performance. The full results and visualizations are available in~Appendix~\ref{app_decorr}.

\section{Conclusion}

Providing explanations for automated predictions carries much more impact, increases transparency, and might even be necessary.
Past work has proposed using attention mechanisms or rationale methods to explain the prediction of a target variable. The former produce noisy explanations, while the latter do not properly capture the multi-faceted nature of useful rationales. Because of the non-probabilistic assignment of words as justifications, rationale methods are prone to suffer from ambiguities and spurious correlations and thus, rely on unrealistic assumptions about the data.

The Multi-Target Masker (MTM) addresses these drawbacks by replacing the binary mask with a probabilistic multi-dimensional mask (one dimension per target), learned in an unsupervised and multi-task learning manner, while jointly predicting all the target variables.

According to comparison with human annotations and automatic evaluation on two real-world datasets, the inferred masks were more accurate and coherent than those that were produced by the state-of-the-art methods.
It is the first technique that delivers both the best explanations and highest accuracy for multiple targets simultaneously.

\newpage
\clearpage
\appendix

\section{Appendices}

\subsection{Topic Words per Aspect}
\label{sec_topic_words}

For each model, we computed the probability distribution of words per aspect by using the induced sub-masks $M_{a_1}, ..., M_{a_A}$ or attention values. Given an aspect $a_i$ and a set of top-$N$ words $\boldsymbol{w_{a_i}^N}$, the Normalized Pointwise Mutual Information~\cite{Bouma2009} coherence score is: \begin{equation}
    \text{NPMI}(\boldsymbol{w_{a_i}^N}) = \sum_{j=2}^{N} \sum_{k=1}^{j-1} \frac{\log \frac{P(w_{a_i}^k, w_{a_i}^j)}{P(w_{a_i}^k)P(w_{a_i}^j)}}{-\log P(w_{a_i}^k, w_{a_i}^j)}
\end{equation}
Top words of coherent topics (i.e., aspects) should share a similar semantic interpretation, and thus interpretability of a topic can be estimated by measuring how many words are not related. For each aspect $a_i$ and word $w$ having been highlighted at least once as belonging to aspect $a_i$, we computed the probability $P(w|a_i)$ on each dataset and sorted them in decreasing order of $P(w|a_i)$. Unsurprisingly, we found that the most common words are stop words such as ``a" and ``it", because masks are mostly word sequences instead of individual words. To gain a better interpretation of the aspect words, we followed the procedure in \cite{McAuley2012}: we first computed the averages across all aspect words for each word $w$ as follows: \begin{equation}
	b_w = \frac{1}{|A|}\sum_{i=1}^{|A|}P(w|a_i)
\end{equation} It represents a general distribution that includes words common to all aspects. The final word distribution per aspect is computed by removing the general distribution as follows:
\begin{equation}
	\hat{P}(w|a_i) = P(w|a_i) - b_w
\end{equation}
After generating the final word distribution per aspect, we picked the top ten words and asked two human annotators to identify intruder words (i.e., words not matching the corresponding aspect). We show in Table~\ref{topics_words_full_beer} and Table~\ref{topics_words_hotel} (and also Table~\ref{topics_words_dec_beer} in Appendix~\ref{app_decorr}) the top ten words for each aspect, where \textcolor{redo}{\textbf{red}} denotes all words identified as unrelated to the aspect by the two annotators. Generally, our model finds better sets of words across the three datasets compared with other methods. Additionally, we observe that the aspects can be easily recovered, given its top words.

\begin{table*}[!htb]
\centering
\begin{tabular}{@{}c@{\hspace*{3mm}}l@{\hspace*{2mm}}l@{}}
\multicolumn{1}{c}{\textbf{}} & \multicolumn{1}{c}{\textbf{Model}} & \multicolumn{1}{c}{\textbf{Top-10 Words}}\\
\toprule

\multirow{4}{*}{\rotatebox{90}{\textit{Appearance}}}
& SAM & \textcolor{redo}{\textbf{nothing}}	beautiful	lager	nice	\textcolor{redo}{\textbf{average}}	macro	lagers	corn	\textcolor{redo}{\textbf{rich}}	gorgeous\\
& MASA & lacing	head	lace	\textcolor{redo}{\textbf{smell}}	amber	retention	beer	nice	carbonation	glass\\
& MAA & head lacing \textcolor{redo}{\textbf{smell}} \textcolor{redo}{\textbf{aroma}} color pours amber glass white retention\\
& MTM (Ours) & head	lacing	\textcolor{redo}{\textbf{smell}}	white	lace	retention	glass	\textcolor{redo}{\textbf{aroma}} tan	thin\\
\cmidrule{1-3}
\multirow{4}{*}{\rotatebox{90}{\textit{Smell}}}
& SAM & faint	\textcolor{redo}{\textbf{nice}}	\textcolor{redo}{\textbf{mild}}	light	slight	complex	good	wonderful	grainy	great\\
& MASA & aroma	hops	nose	chocolate	caramel	malt	citrus	fruit	smell	fruits\\
& MAA & \textcolor{redo}{\textbf{taste}} hints hint \textcolor{redo}{\textbf{lots}} \textcolor{redo}{\textbf{t-}} \textcolor{redo}{\textbf{starts}} blend mix \textcolor{redo}{\textbf{upfront}} malts\\
& MTM (Ours) & \textcolor{redo}{\textbf{taste}} 	malt	aroma	hops	sweet	citrus	caramel	nose	malts	chocolate \\
\cmidrule{1-3}
\multirow{4}{*}{\rotatebox{90}{\textit{Palate}}} 
& SAM & thin	\textcolor{redo}{\textbf{bad}}	light	watery	creamy	silky	medium	body	smooth	\textcolor{redo}{\textbf{perfect}}\\
& MASA & smooth	light	medium	thin	creamy	\textcolor{redo}{\textbf{bad}}	watery	\textcolor{redo}{\textbf{full}}	crisp	\textcolor{redo}{\textbf{clean}}\\
& MAA & good beer carbonation smooth \textcolor{redo}{\textbf{drinkable}} medium bodied \textcolor{redo}{\textbf{nice}} body \textcolor{redo}{\textbf{overall}}\\
& MTM (Ours) & carbonation	medium	mouthfeel	body	smooth	bodied \textcolor{redo}{\textbf{drinkability}}	creamy	light	\textcolor{redo}{\textbf{overall}} \\
\cmidrule{1-3}
\multirow{4}{*}{\rotatebox{90}{\textit{Taste}}}
& SAM & \textcolor{redo}{\textbf{decent}}	great	complex	delicious	tasty	favorite	\textcolor{redo}{\textbf{pretty}}	sweet	\textcolor{redo}{\textbf{well}}	\textcolor{redo}{\textbf{best}}\\
& MASA & good	\textcolor{redo}{\textbf{drinkable}}	\textcolor{redo}{\textbf{nice}}	tasty	great	enjoyable	\textcolor{redo}{\textbf{decent}}	\textcolor{redo}{\textbf{solid}}	balanced	\textcolor{redo}{\textbf{average}}\\
& MAA & malt hops flavor hop flavors caramel malts bitterness bit chocolate\\
& MTM (Ours) & malt	sweet	hops	flavor	bitterness	finish	chocolate	bitter	caramel	sweetness\\
\end{tabular}

\caption{\label{topics_words_full_beer}Top ten words for each aspect from the \textbf{\textit{Beer}} dataset, learned by various models. \textcolor{redo}{\textbf{Red}} denotes intruders according to two annotators. Found words are generally noisier due to the high correlation between \textit{Taste} and other aspects. However, \textit{MTM} provides better results than other methods.}
\end{table*}

\begin{table*}[!htb]
\centering
\begin{tabular}{@{}c@{\hspace*{3mm}}l@{\hspace*{2mm}}l@{}}
\multicolumn{1}{c}{\textbf{}} & \multicolumn{1}{c}{\textbf{Model}} & \multicolumn{1}{c}{\textbf{Top-10 Words}}\\
\toprule
\multirow{4}{*}{\rotatebox{90}{\textit{Service}}}
& SAM & staff service friendly nice told helpful good great lovely manager\\
& MASA & friendly helpful told rude nice good pleasant asked enjoyed worst\\
& MAA & staff service helpful friendly nice good rude excellent great desk\\
& MTM (Ours) & staff friendly service desk helpful manager reception told rude asked \\
\cmidrule{1-3}
\multirow{4}{*}{\rotatebox{90}{\textit{Cleanliness}}}
& SAM & clean cleaned dirty toilet smell cleaning sheets comfortable nice hair\\
& MASA & clean dirty cleaning spotless stains cleaned cleanliness mold filthy bugs \\
& MAA & clean dirty cleaned filthy stained well spotless carpet sheets stains\\
& MTM (Ours) & clean dirty bathroom room bed cleaned sheets smell carpet toilet\\
\cmidrule{1-3}
\multirow{4}{*}{\rotatebox{90}{\textit{Value}}}
& SAM & good stay great well \textcolor{redo}{\textbf{dirty}} recommend worth definitely \textcolor{redo}{\textbf{friendly}} charged\\
& MASA & great good poor excellent terrible awful \textcolor{redo}{\textbf{dirty}} horrible \textcolor{redo}{\textbf{disgusting}} \textcolor{redo}{\textbf{comfortable}}\\
& MAA &\textcolor{redo}{\textbf{night}} \textcolor{redo}{\textbf{stayed}} stay \textcolor{redo}{\textbf{nights}} \textcolor{redo}{\textbf{2}} \textcolor{redo}{\textbf{day}} price \textcolor{redo}{\textbf{water}} \textcolor{redo}{\textbf{4}} \textcolor{redo}{\textbf{3}}\\
& MTM (Ours) & good price expensive paid cheap worth better pay overall disappointed\\
\cmidrule{1-3}
\multirow{4}{*}{\rotatebox{90}{\textit{Location}}}
& SAM & location close far place walking \textcolor{redo}{\textbf{definitely}} located \textcolor{redo}{\textbf{stay}} short view\\
& MASA & location beach walk hotel town located restaurants walking close taxi\\
& MAA & location hotel place located close far area beach view situated\\
& MTM (Ours) & location great area walk beach hotel town close city street\\
\cmidrule{1-3}
\multirow{4}{*}{\rotatebox{90}{\textit{Room}}}
& SAM &\textcolor{redo}{\textbf{dirty} \textbf{clean}} small best comfortable large worst modern \textcolor{redo}{\textbf{smell}} spacious\\
& MASA & comfortable small spacious nice large dated well tiny modern basic \\
& MAA & room rooms bathroom bed spacious small beds large shower modern\\
& MTM (Ours)& comfortable room small spacious nice modern rooms large tiny walls\\
\end{tabular}

\caption{\label{topics_words_hotel}Top ten words for each aspect from the \textbf{\textit{Hotel}} dataset, learned by various models. \textcolor{redo}{\textbf{Red}} denotes intruders according to human annotators. Besides \textit{SAM}, all methods find similar words for most aspects except the aspect \textit{Value}. The top words of \textit{MTM} do not contain any intruder.}
\end{table*}

\begin{table*}[ht]
\centering
\begin{tabular}{@{}l@{\hspace*{2mm}}c@{\hspace*{2mm}}c@{\hspace*{2mm}}c@{}}
& & & \textbf{Decorrelated}\\
\multicolumn{1}{c}{\bf Dataset} & \multicolumn{1}{c}{\bf Beer}  & \multicolumn{1}{c}{\bf Hotel} & \multicolumn{1}{c}{\bf Beer} \\
\toprule
Number of reviews& $1,586,259$ & $140,000$& $280,000$   \\
Average word-length of review& $147.1 \pm 79.7$  & $188.3 \pm 50.0$ & $157.5 \pm 84.3$ \\
Average sentence-length of review& $10.3 \pm 5.4$  & $10.4 \pm 4.4$ & $11.0 \pm 5.7$ \\
Number of aspects& $4$ & $5$ & $3$ \\
Average ratio of $\oplus$ over $\ominus$ reviews per aspect& $12.89$ & $1.02$ & $3.29$ \\
Average correlation between aspects& $\mathbf{71.8\%}$ & $\mathbf{63.0\%}$ & $\mathbf{27.2\%}$ \\
Max correlation between two aspects& $\mathbf{73.4\%}$ & $\mathbf{86.5\%}$ & $\mathbf{29.8\%}$ \\
\end{tabular}

\caption{\label{app_dataset_description}Statistics of the multi-aspect review datasets. \textit{Beer} and \textit{Hotel} represent real-world beer and hotel reviews, respectively. \textit{Decorrelated Beer} is a subset of the \textit{Beer} dataset with a low-correlation assumption between aspect ratings, leading to a more straightforward and unrealistic dataset.}
\end{table*}

\begin{table*}[t]
\centering
\begin{tabular}{@{}l>{}ll@{}>{}c@{}>{}c@{}>{}c@{\hspace*{2mm}}>{}c@{\hspace*{2mm}}>{}c@{\hspace*{2mm}}>{}c@{}}
& & & & \multicolumn{4}{c}{\textbf{F1 Score}}\\
\cmidrule(lr){5-8}
\multicolumn{1}{c}{\bf Interp.} & \multicolumn{2}{c}{\bf Model} & \multicolumn{1}{c}{\bf Params} & \multicolumn{1}{c}{\bf Macro} & \multicolumn{1}{c}{\boldmath$A_1$} & \multicolumn{1}{c}{\boldmath$A_2$} & \multicolumn{1}{c}{\boldmath$A_3$}\\
\toprule
\multicolumn{1}{c}{\multirow{2}{*}{\parbox{1.cm}{\centering None}}}
& SENT & Sentiment Majority & $426k$ & $68.89$ & $67.48$ & $73.49$ & $65.69$\\
& BASE & $\text{Emb}_{200}$ + $\text{Enc}_\textsubscript{CNN}$ + \text{Clf} & $173k$ & $78.23$ & $78.38$ & $80.86$ & $75.47$\\
\midrule
\multicolumn{1}{c}{\multirow{2}{*}{\parbox{1.cm}{\centering Coarse-grained}}}
& \multirow{2}{*}{SAA} & $\text{Emb}_{200}$ + $\text{Enc}_\textsubscript{CNN}$ + $\text{A}_\textsubscript{Shared}$ + \text{Clf} & $196k$ & $78.19$ & $77.43$ & $80.96$ & $76.16$\\
& & $\text{Emb}_{200}$ + $\text{Enc}_\textsubscript{LSTM}$ + $\text{A}_\textsubscript{Shared}$ + \text{Clf} & $186k$ & $78.16$ & $75.88$ & $81.25$ & $77.36$\\
\midrule
\multicolumn{1}{c}{\multirow{6}{*}{\parbox{1.cm}{\centering Fine-grained}}}
& NB-SVM & \cite{wang-manning-2012-baselines} & $3 \cdot 426k$ & $74.60$ & $73.50$ & $77.32$ & $72.99$\\
& SAM & \cite{lei-etal-2016-rationalizing} & $3 \cdot 644k$ & $77.06$ & $77.36$ & $78.99$ & $74.83$\\
& MASA & $\text{Emb}_{200}$ + $\text{Enc}_\textsubscript{LSTM}$ + $\text{A}^\textsubscript{Sparse}_\textsubscript{Aspect-wise}$ + \text{Clf} & $458k$ & $78.82$ & $77.35$ & $81.65$ & $77.47$\\
& MAA & $\text{Emb}_{200}$ + $\text{Enc}_\textsubscript{LSTM}$ + $\text{A}_\textsubscript{Aspect-wise}$ + \text{Clf} & $458k$ & $78.96$ & $78.54$ & $81.56$ & $76.79$\\
\cdashlinelr{2-8}
& MTM\textsuperscript{} & $\text{Emb}_{200}$ + \text{Masker} + $\text{Enc}_\textsubscript{CNN}$ + \text{Clf} (Ours) & $274k$ & $79.32$ & $78.58$ & $81.71$ & $77.66$\\
& \textbf{MTM\textsuperscript{C}} & \textbf{$\text{Emb}_{200+4}$ +  $\text{Enc}_\textsubscript{CNN}$ + \text{Clf} (Ours)} & $175k$ & $\mathbf{79.66}$ & $\mathbf{78.74}$ & $\mathbf{82.02}$ & $\mathbf{78.22}$ \\
\end{tabular}

\caption{\label{app_perfs_small_beer}Performance of the multi-aspect sentiment classification task for the \textbf{\textit{decorrelated}} \textit{Beer} dataset.}
\end{table*}

\begin{table*}[!htb]
\centering
\begin{tabular}{@{}c@{\hspace*{3mm}}l@{\hspace*{2mm}}l@{}}
\multicolumn{1}{c}{\textbf{}} & \multicolumn{1}{c}{\textbf{Model}} & \multicolumn{1}{c}{\textbf{Top-10 Words}}\\
\toprule

\multirow{4}{*}{\rotatebox{90}{\textit{Appearance}}}
& SAM  & head color white brown dark lacing \textcolor{redo}{\textbf{pours}} amber clear black\\
& MASA & head	lacing	lace	retention	glass	foam	color	amber	yellow	cloudy\\
& MAA & nice dark amber \textcolor{redo}{\textbf{pours}} black hazy brown \textcolor{redo}{\textbf{great}} cloudy clear\\
& MTM (Ours) & head color lacing white brown clear amber glass black retention \\
\cmidrule{1-3}
\multirow{4}{*}{\rotatebox{90}{\textit{Smell}}}
& SAM & sweet malt hops coffee chocolate citrus hop strong smell aroma\\
& MASA & smell	aroma	nose	smells	sweet	aromas	scent	hops	malty	roasted\\
& MAA &\textcolor{redo}{\textbf{taste}} smell aroma sweet chocolate \textcolor{redo}{\textbf{lacing}} malt roasted hops nose\\
& MTM (Ours) & smell aroma nose smells sweet malt citrus chocolate caramel aromas\\
\cmidrule{1-3}
\multirow{4}{*}{\rotatebox{90}{\textit{Palate}}} 
& SAM & mouthfeel smooth medium carbonation bodied watery body thin creamy \textcolor{redo}{\textbf{full}}\\
& MASA & mouthfeel	medium	smooth	body	\textcolor{redo}{\textbf{nice}}	m-	feel	bodied	mouth	\textcolor{redo}{\textbf{beer}}\\
& MAA & carbonation mouthfeel medium \textcolor{redo}{\textbf{overall}} smooth finish body \textcolor{redo}{\textbf{drinkability}} bodied watery\\
& MTM (Ours)& mouthfeel carbonation medium smooth body bodied \textcolor{redo}{\textbf{drinkability}} good mouth thin\\
\end{tabular}

\caption{\label{topics_words_dec_beer}Top ten words for each aspect from the \textbf{\textit{decorrelated}} \textit{Beer} dataset, learned by various models. \textcolor{redo}{\textbf{Red}} denotes intruders according to two annotators. For the three aspects, MTM has only one word considered as an intruder, followed by MASA with SAM (two) and MAA (six).}
\end{table*}

\subsection{Results Decorrelated Beer Dataset}
\label{app_decorr}

\begin{figure*}[!htb]
\centering
\begin{tabular}{@{}c@{\makebox[0.25cm]{ }}c@{}}
    \includegraphics[width=0.475\textwidth]{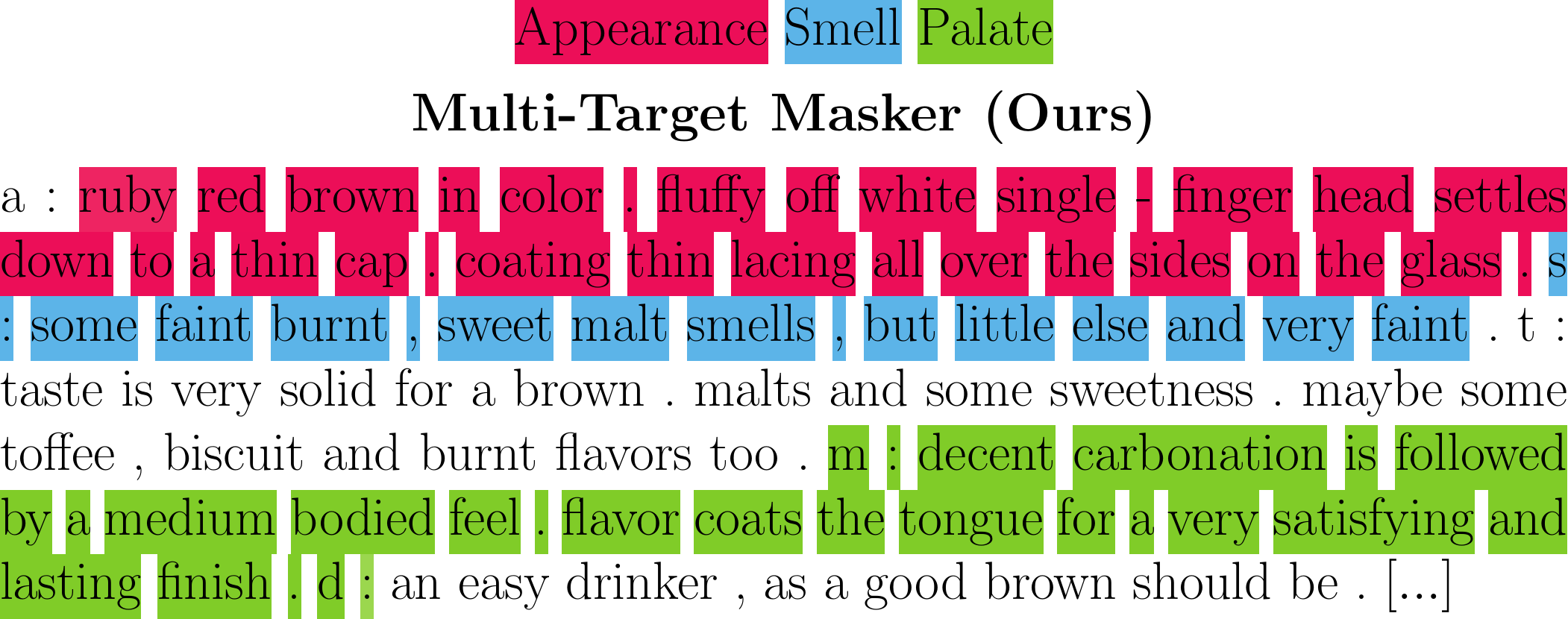} & 
    \includegraphics[width=0.475\textwidth]{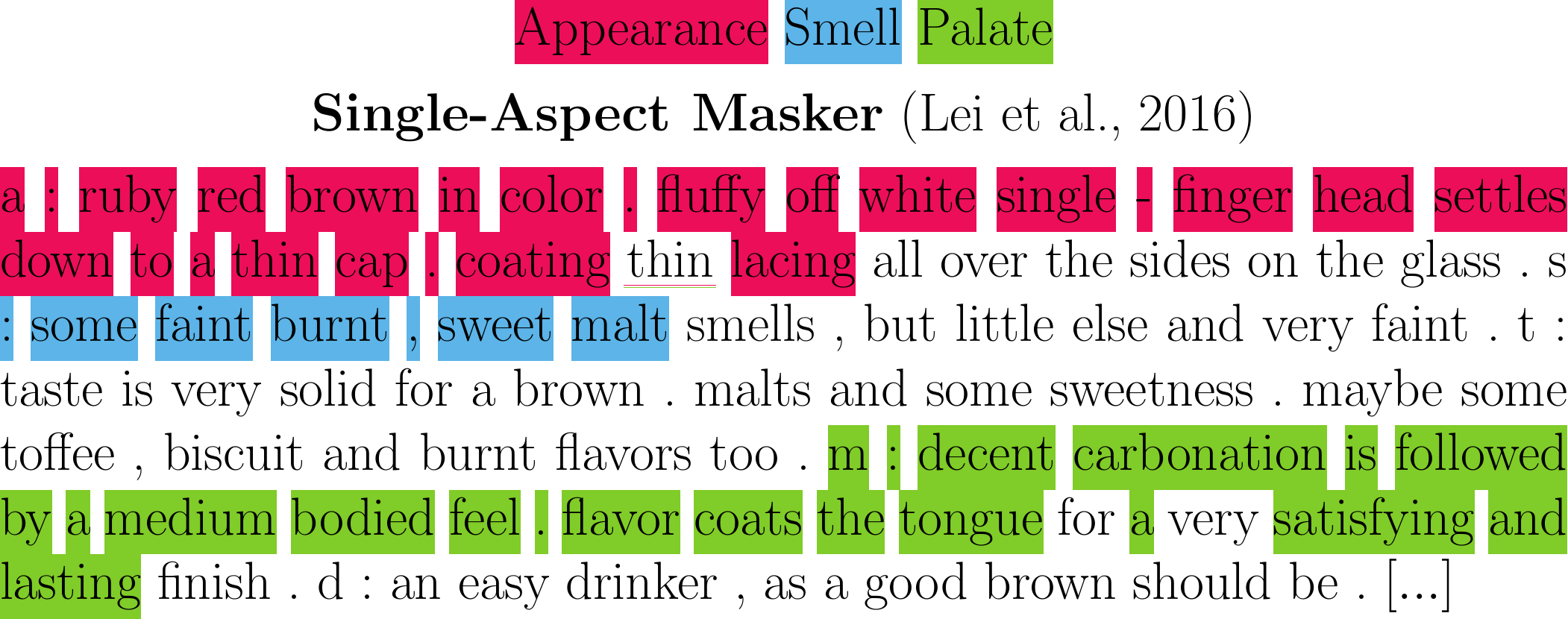} \\\\
    \includegraphics[width=0.475\textwidth]{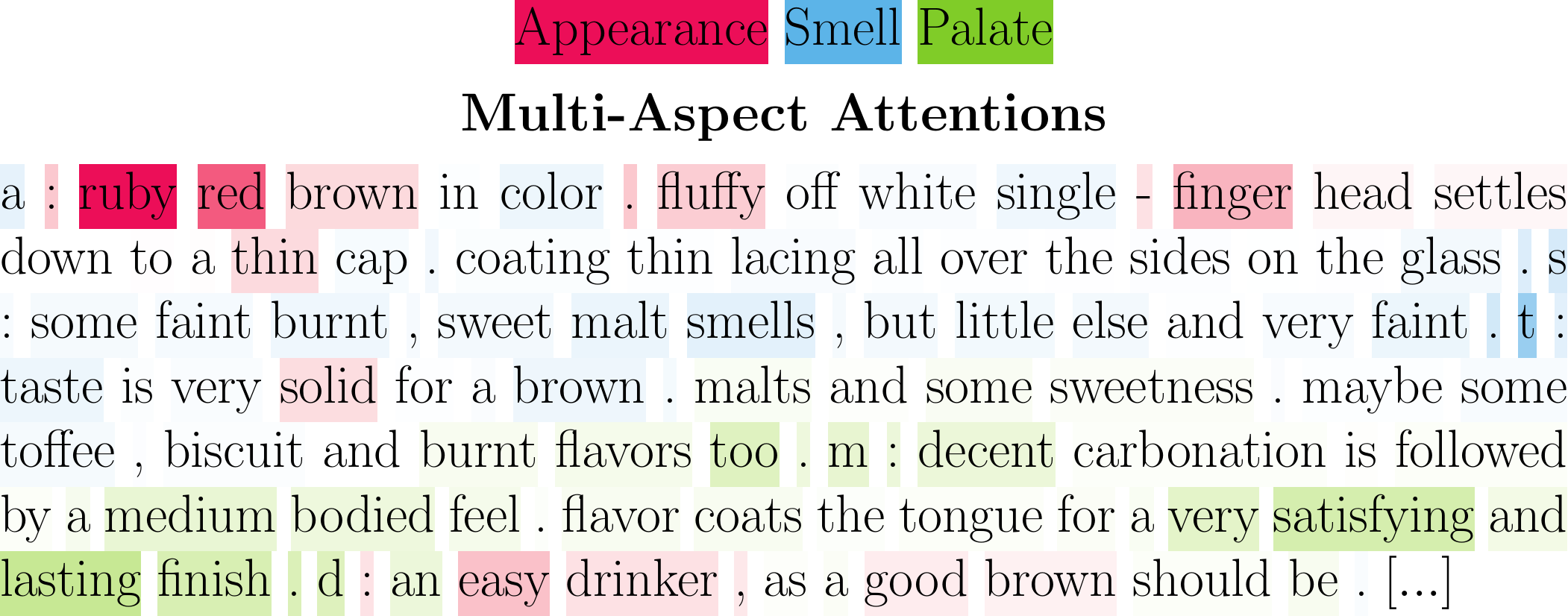} &
    \includegraphics[width=0.475\textwidth]{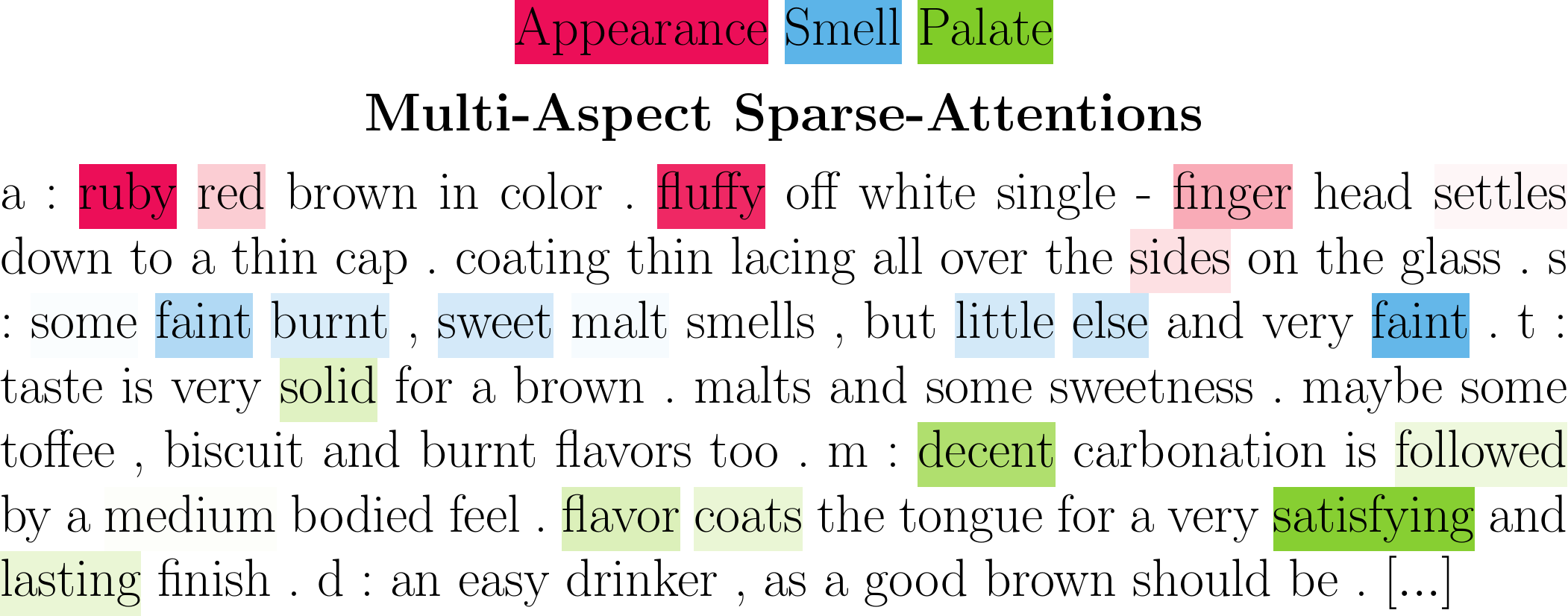}\\\\
\end{tabular}

\caption{\label{sample_beer_0}A sample review from the \textbf{\textit{decorrelated}} \textit{Beer} dataset, with computed masks from different methods. Our model \textit{MTM} highlights all the words corresponding to the aspects. \textit{SAM} only highlights the most crucial information, but some words are missing out and one is ambiguous. \textit{MAA} and \textit{MASA} fail to identify most of the words related to the aspect \textit{Appearance}, and only a few words have high confidence, resulting in noisy labeling. Additionally, \textit{MAA} considers words belonging to the aspect \textit{Taste} whereas this dataset does not include it in the aspect set (because it has a high correlation with other rating scores).}
\end{figure*}

\begin{figure*}[!htb]

\centering
\begin{tabular}{@{}c@{\makebox[0.25cm]{ }}c@{}}
    \includegraphics[width=0.475\textwidth]{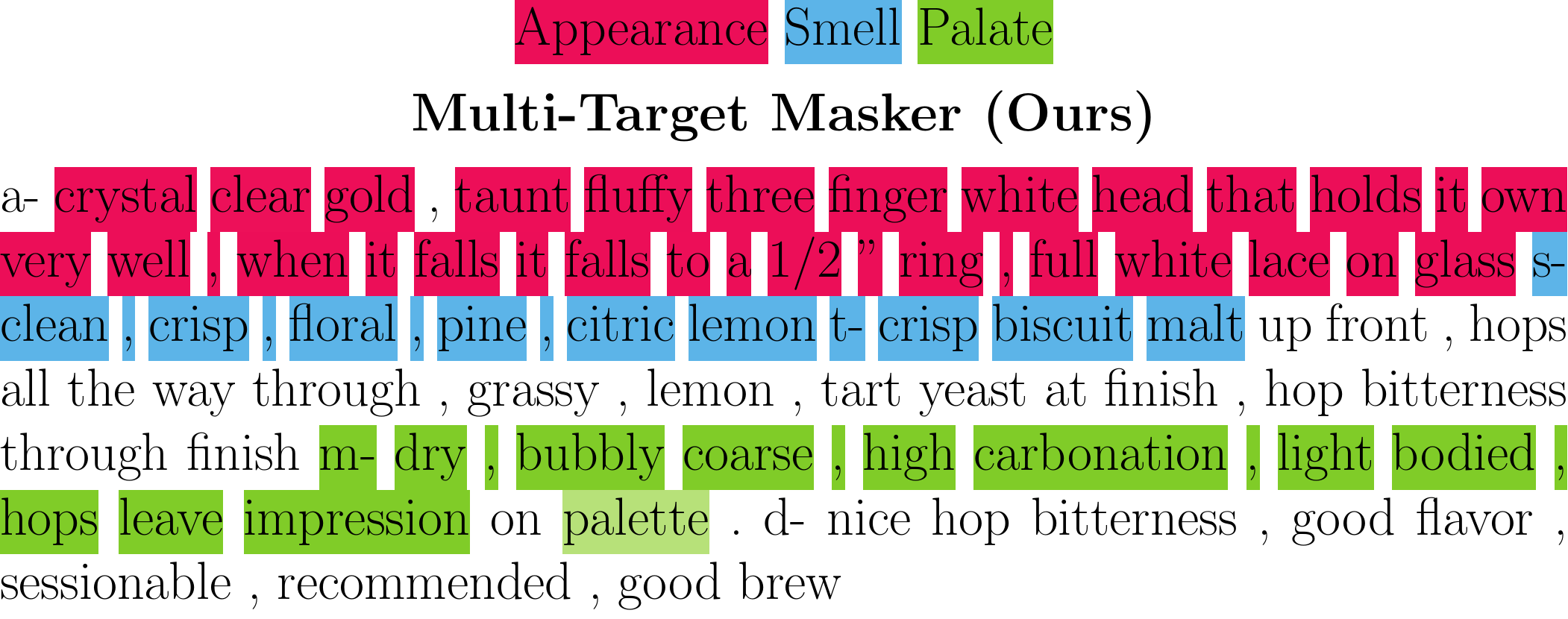} & 
    \includegraphics[width=0.475\textwidth]{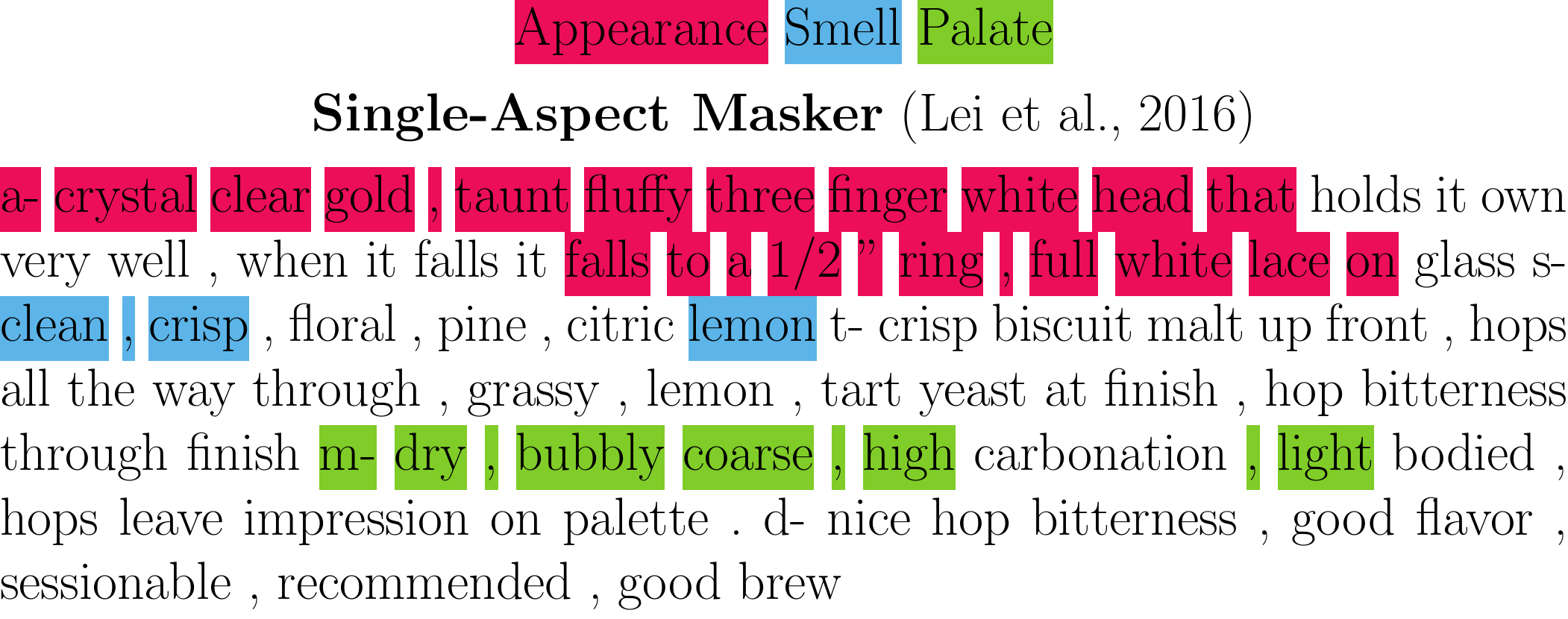} \\\\
    \includegraphics[width=0.475\textwidth]{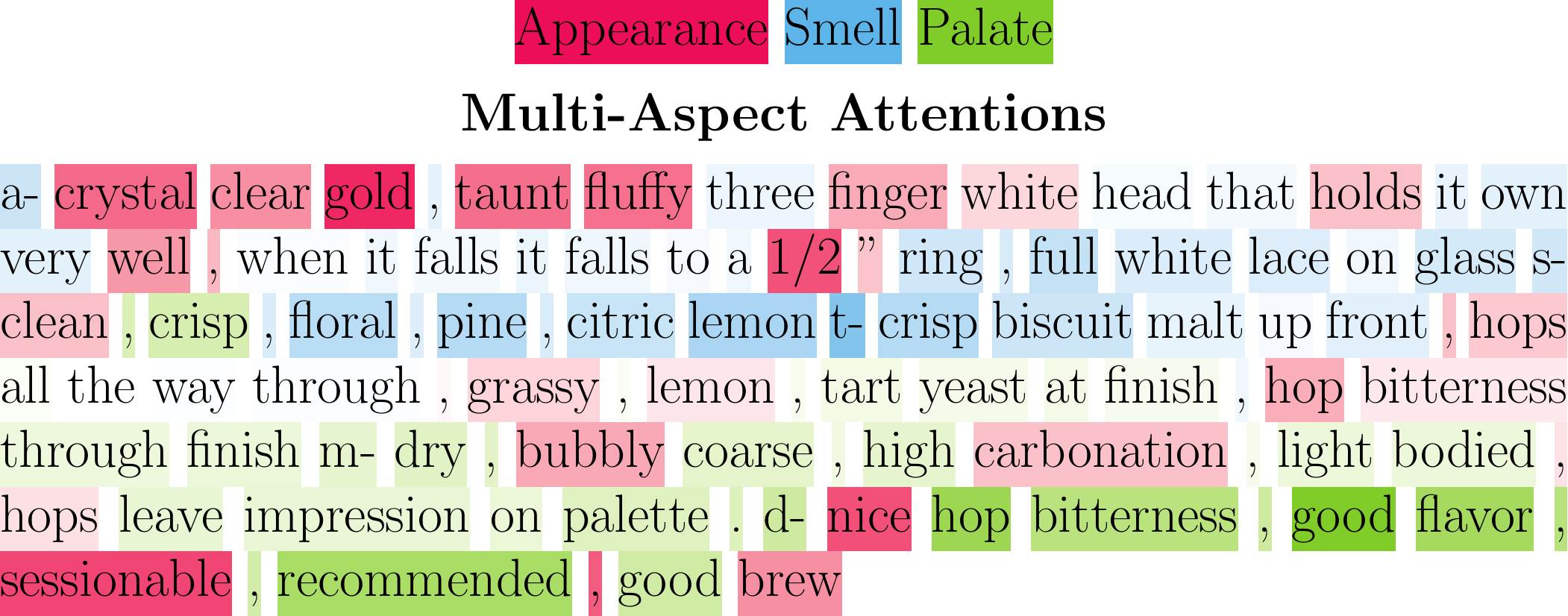} &
    \includegraphics[width=0.475\textwidth]{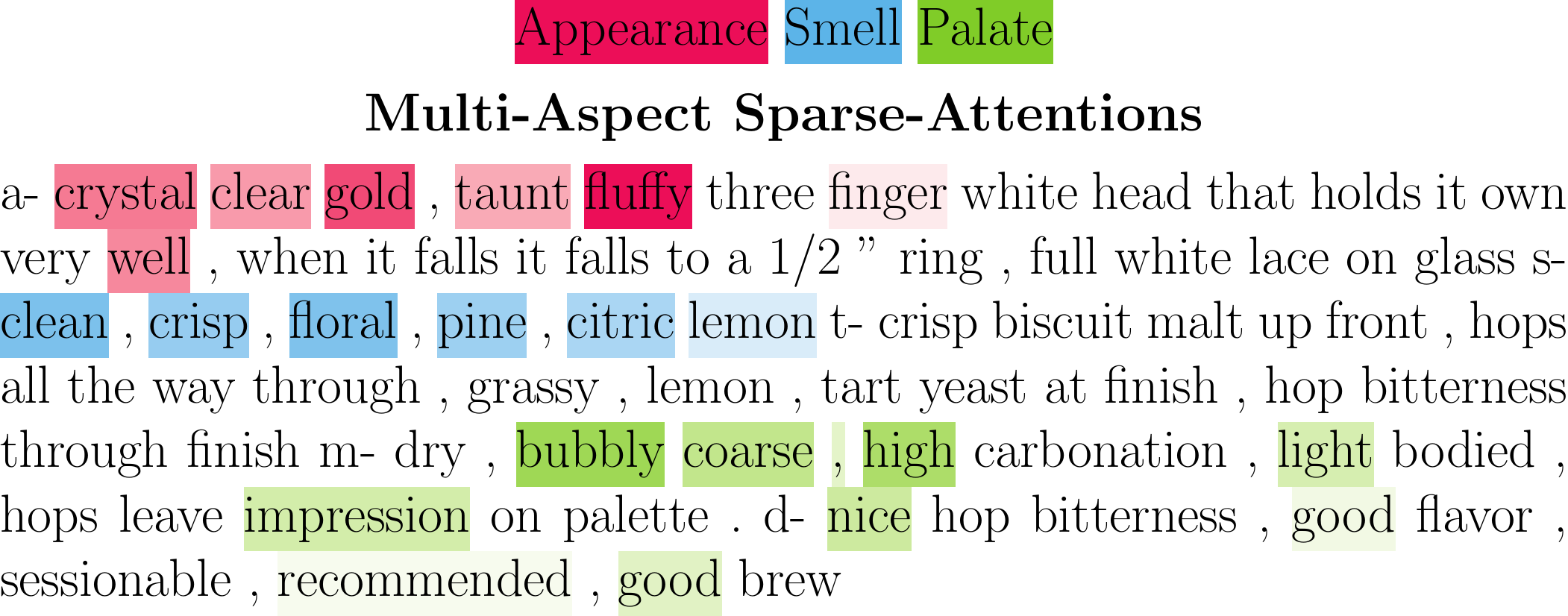}\\\\
\end{tabular}

\caption{\label{sample_beer_1}A sample review from the \textbf{\textit{decorrelated}} \textit{Beer} dataset, with computed masks from different methods. \textit{MTM} finds the exact parts corresponding to the aspect \textit{Appearance} and \textit{Palate} while covering most of the aspect \textit{Smell}. \textit{SAM} identifies key-information without any ambiguity, but lacks coverage. \textit{MAA} highlights confidently nearly all the words while having some noise for the aspect \textit{Appearance}. \textit{MASA} selects confidently only most predictive words.}
\end{figure*}

We provide additional details of Section~\ref{sec_decorr}. Table~\ref{app_dataset_description} presents descriptive statistics of \textit{Beer} and \textit{Hotel} datasets with the \textit{decorrelated} subset of beer reviews from \cite{lei-etal-2016-rationalizing,li2016understanding,chang2019game,chang2020invariant}. The results of the multi-aspect sentiment classification experiment are shown in Table~\ref{app_perfs_small_beer}. Samples are available in Figure~\ref{sample_beer_0} and Figure~\ref{sample_beer_1}. Table~\ref{topics_words_dec_beer} contains the results of the intruder task.

\subsection{Visualization of the Multi-Dimensional Facets of Reviews}
\label{samples_with_visualization}

\begin{figure*}[!htb]
\centering
\begin{tabular}{@{}c@{\makebox[0.25cm]{ }}c@{}}
    \includegraphics[width=0.475\textwidth]{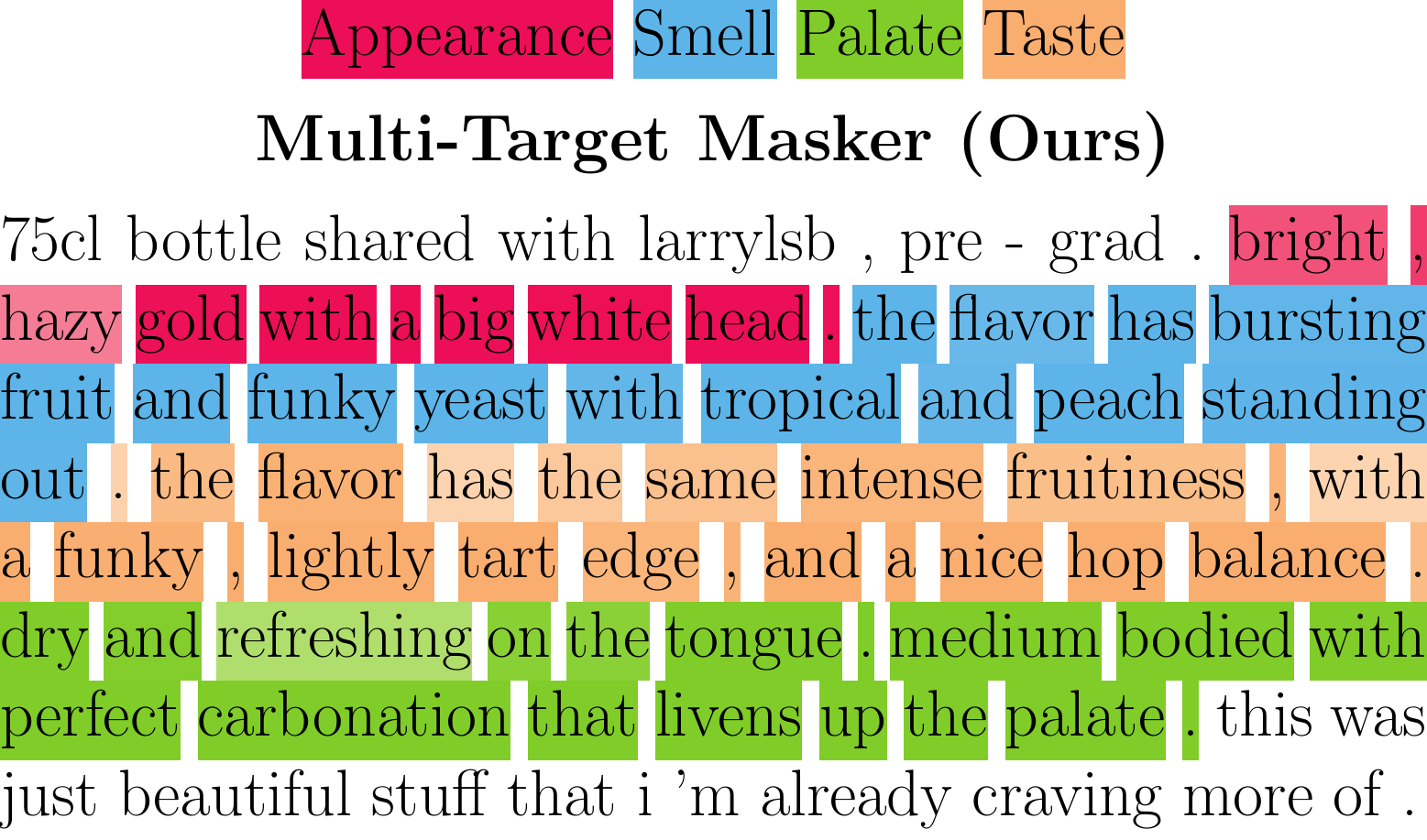} & 
    \includegraphics[width=0.475\textwidth]{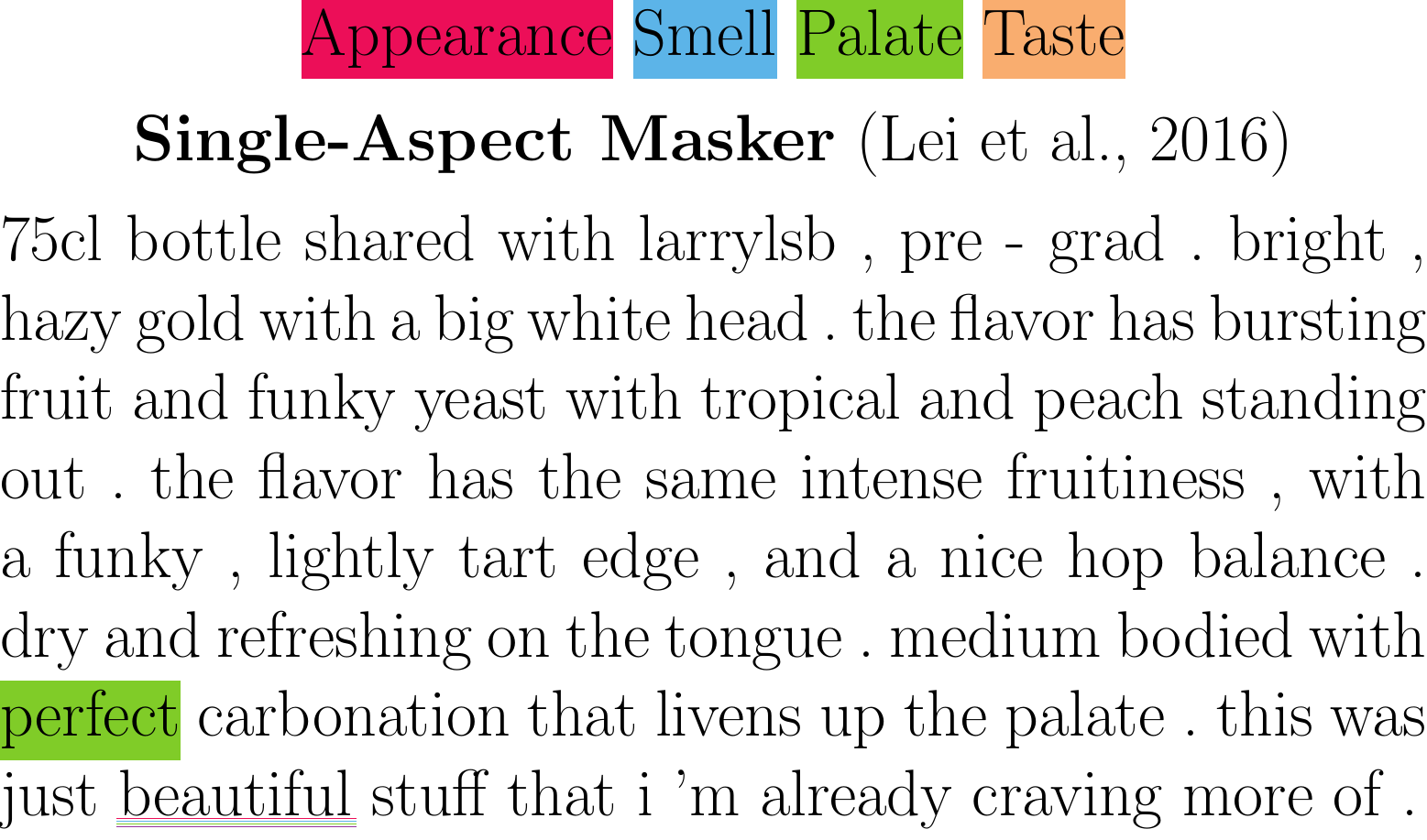} \\\\
    \includegraphics[width=0.475\textwidth]{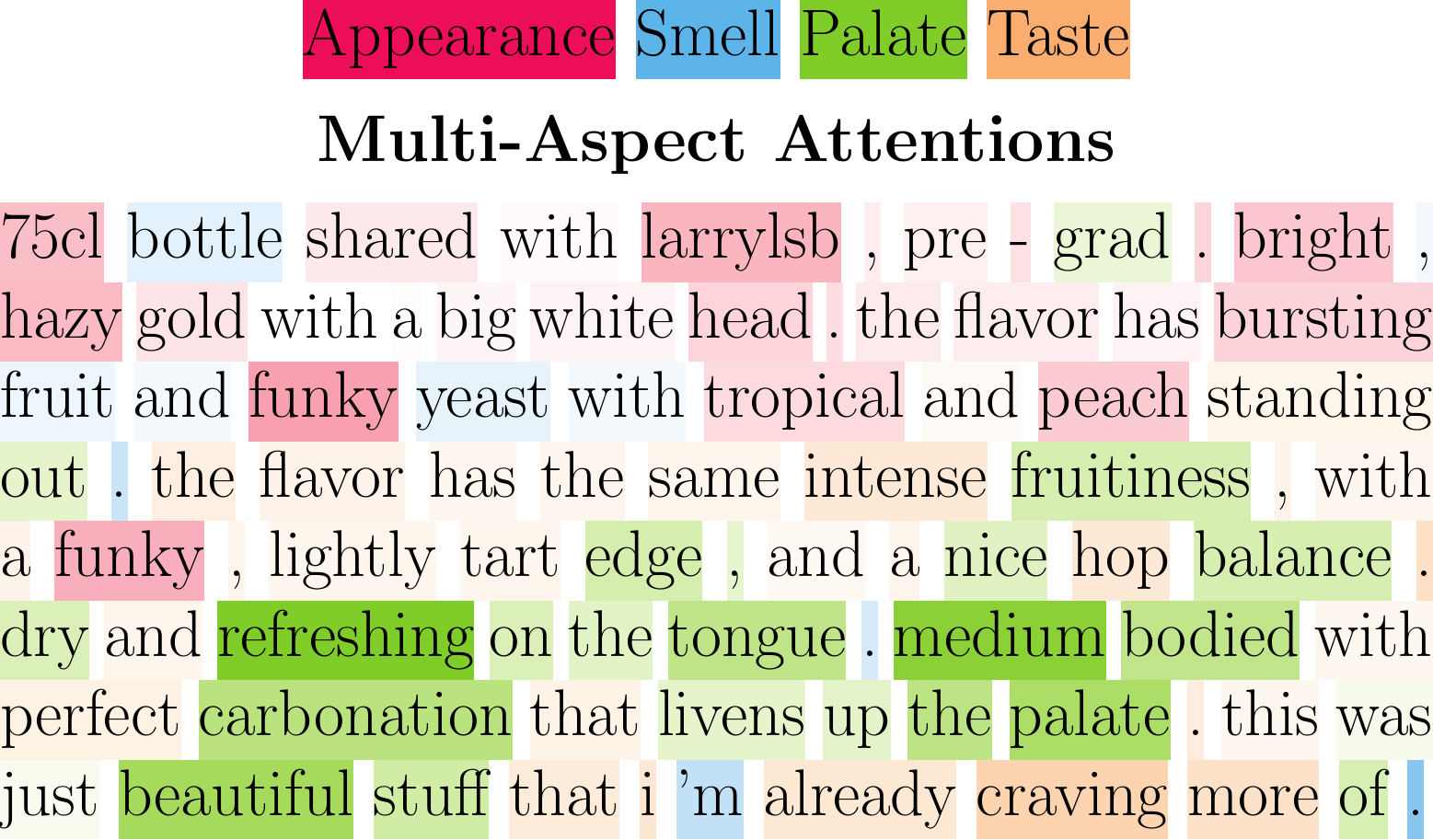} &
    \includegraphics[width=0.475\textwidth]{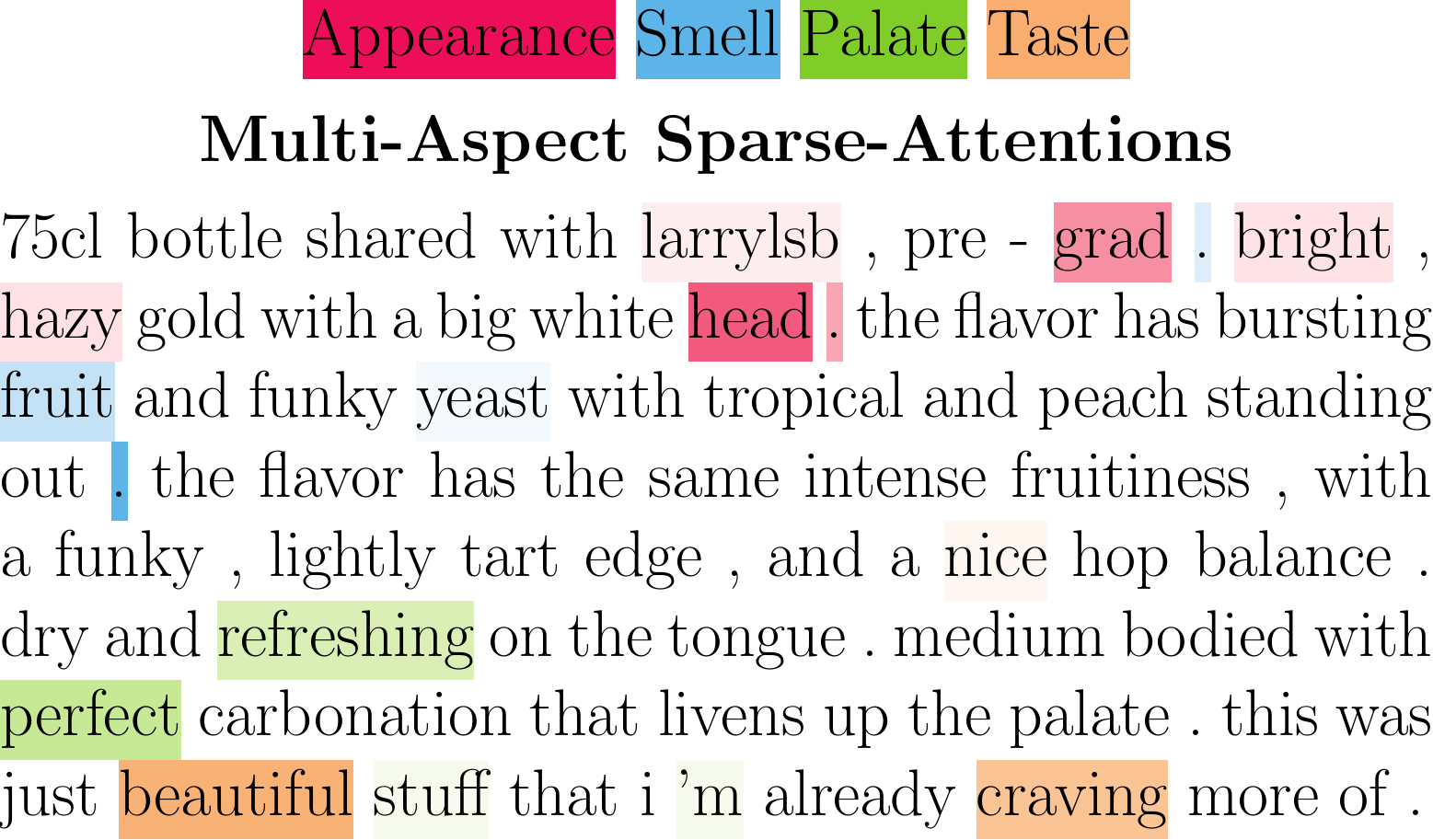}\\\\
\end{tabular}

\caption{\label{sample_full_beer_0}A sample review from the \textit{Beer} dataset, with computed masks from different methods. \textit{MTM} achieves near-perfect annotations, while \textit{SAM} highlights only two words where one is ambiguous with respect to the four aspects. \textit{MAA} mixes between the aspect \textit{Appearance} and \textit{Smell}. \textit{MASA} identifies some words but lacks coverage.}
\end{figure*}

\begin{figure*}[!htb]

\centering
\begin{tabular}{@{}c@{\makebox[0.25cm]{ }}c@{}}
     \includegraphics[width=0.475\textwidth,height=10cm]{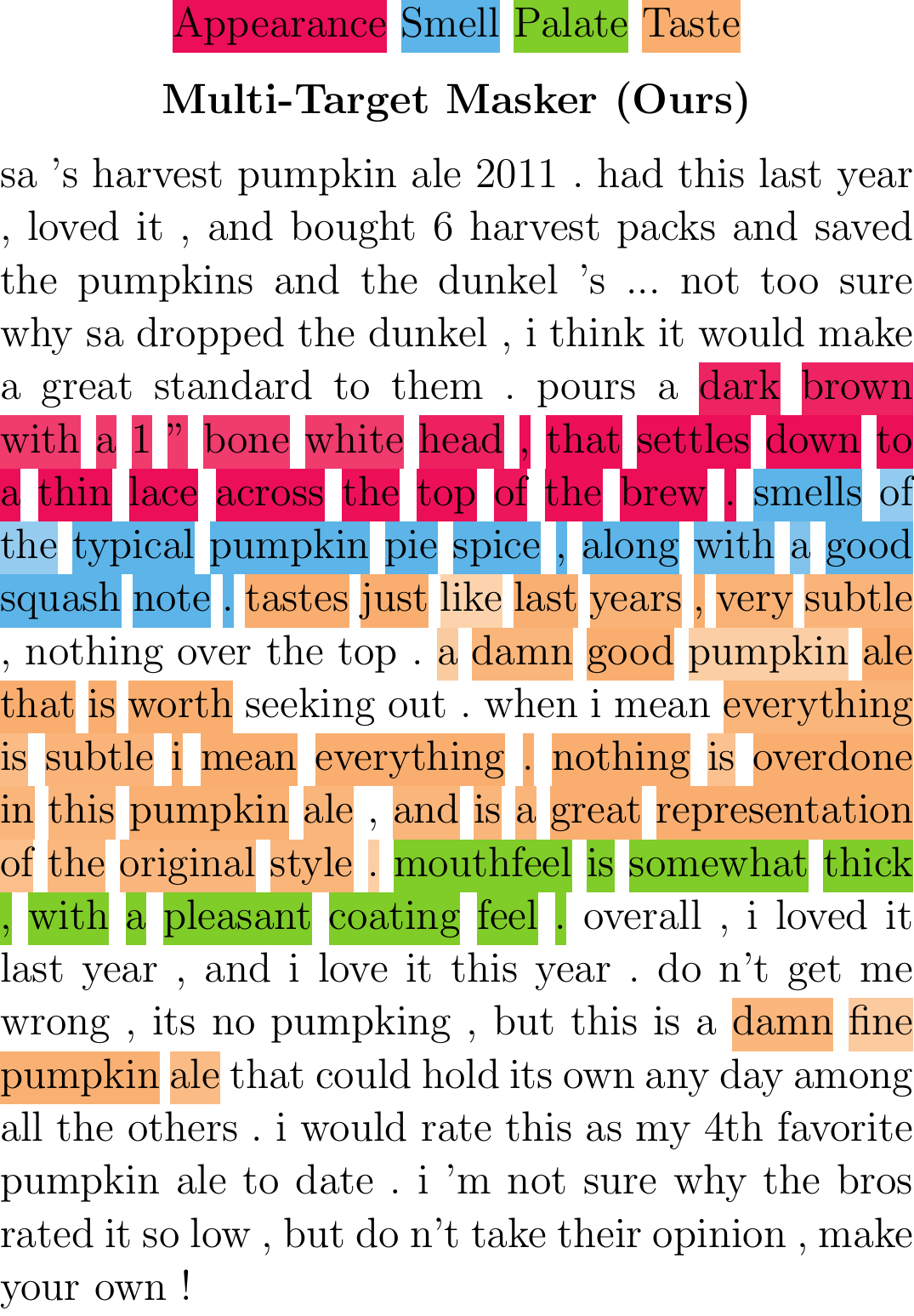} & 
     \includegraphics[width=0.475\textwidth,height=10cm]{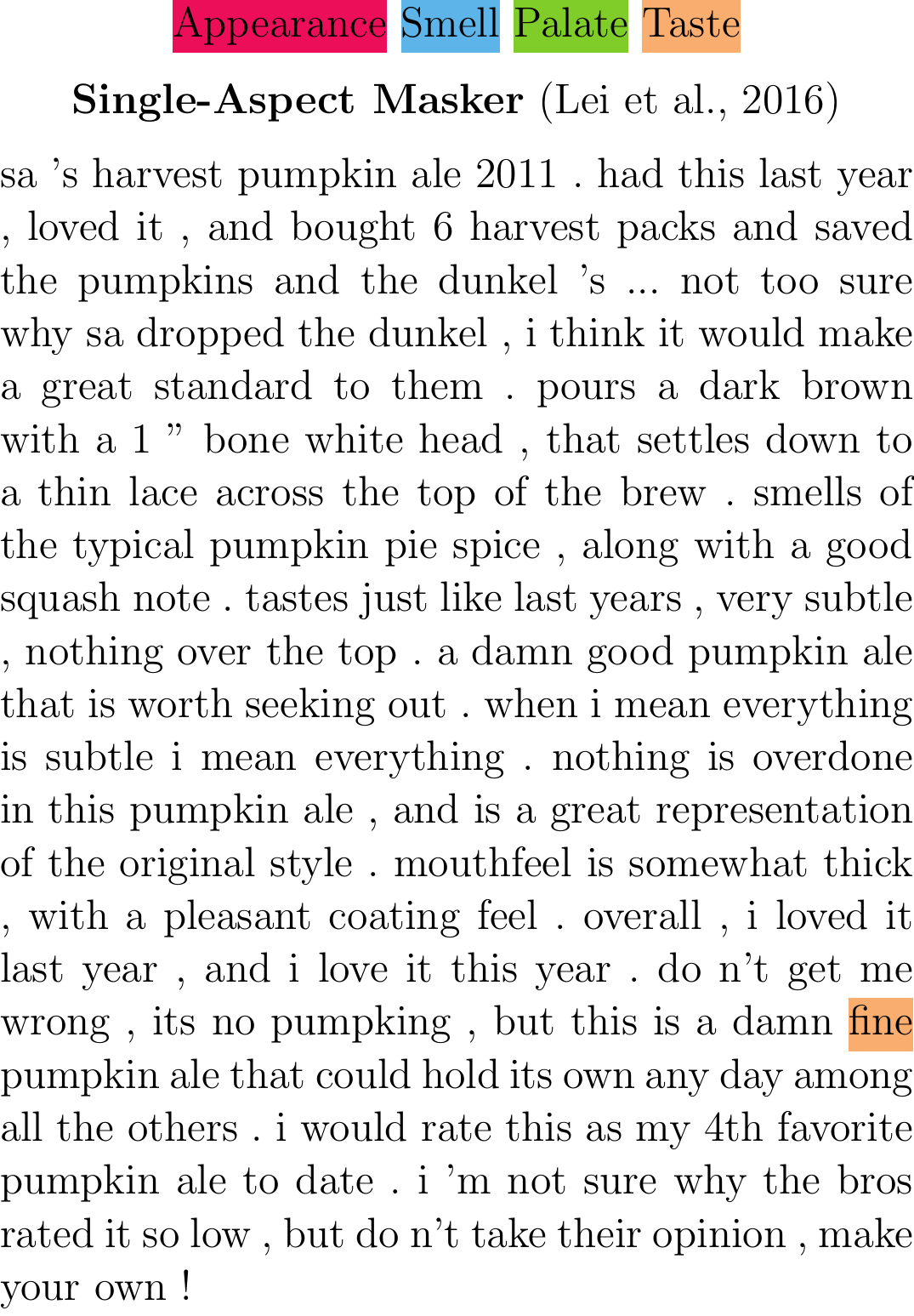} \\\\
     \includegraphics[width=0.475\textwidth,height=10cm]{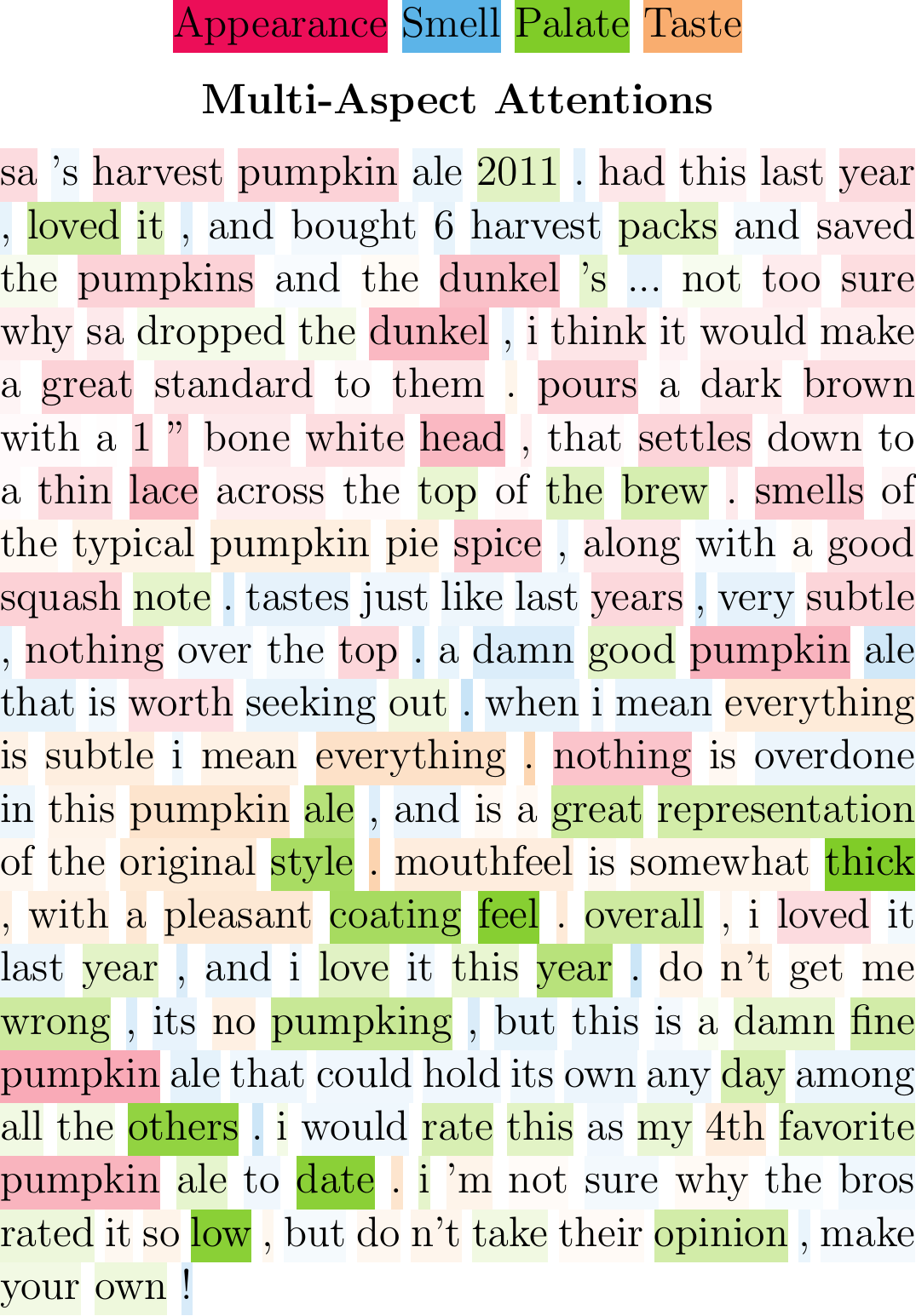} & 
     \includegraphics[width=0.475\textwidth,height=10cm]{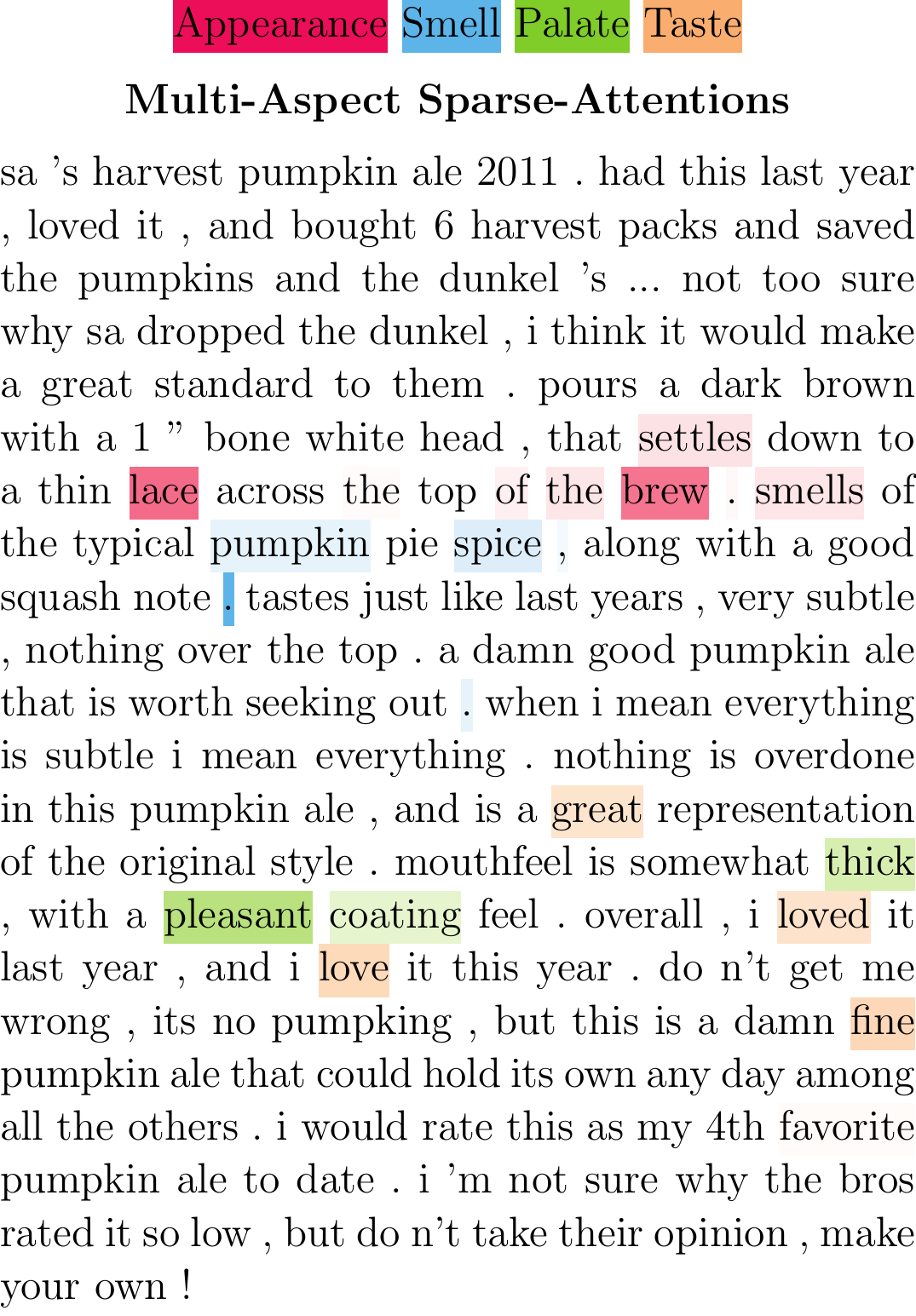}\\
\end{tabular}

\caption{\label{sample_full_beer_1}A sample review from the \textit{Beer} dataset, with computed masks from different methods. \textit{MTM} can accurately identify what parts of the review describe each aspect. \textit{MAA} provides very noisy labels due to the high imbalance and correlation between aspects, while \textit{MASA} highlights only a few important words. We can see that \textit{SAM} is confused and performs a poor selection.}
\end{figure*}

\begin{figure*}[!htb]
\centering
\begin{tabular}{@{}c@{\makebox[0.25cm]{ }}c@{}}
     \includegraphics[width=0.475\textwidth]{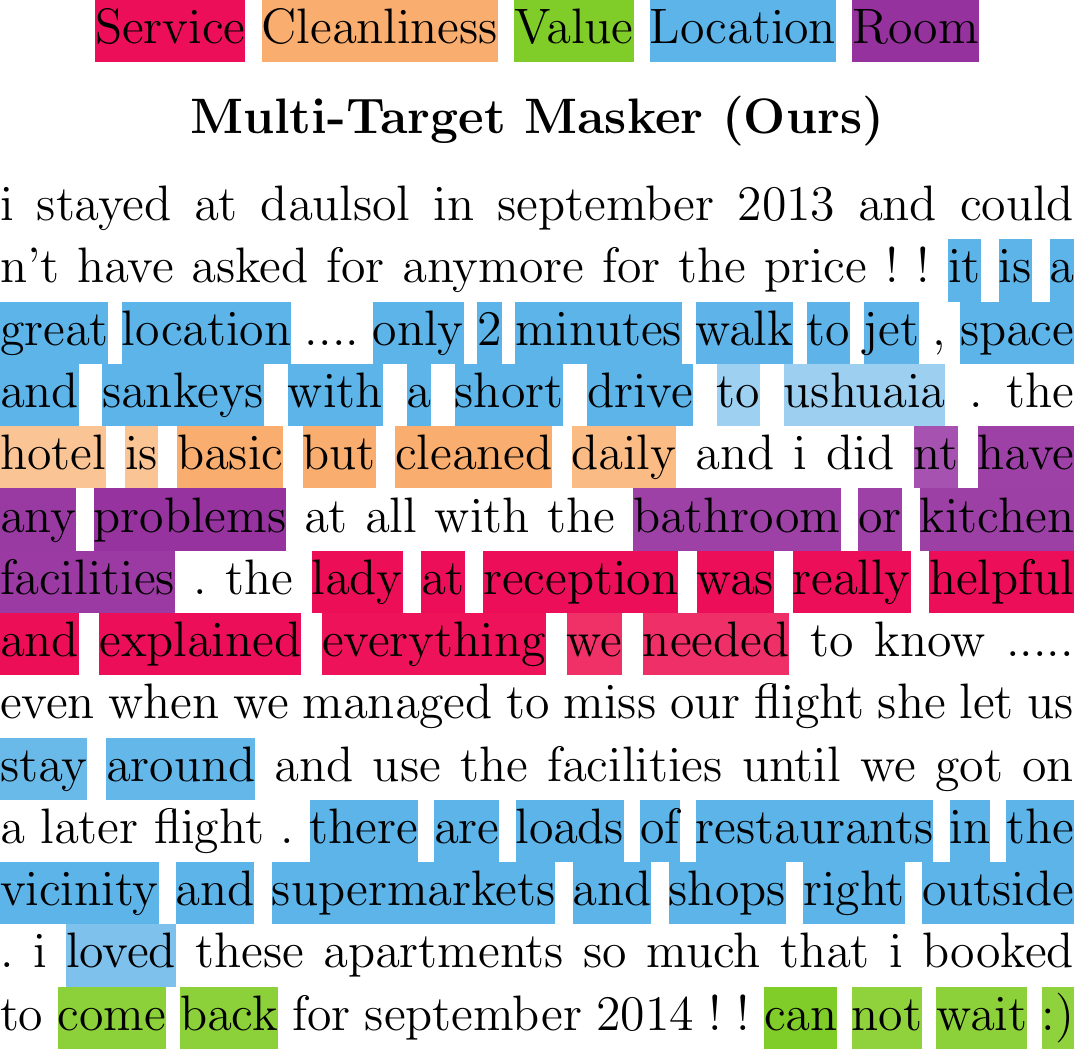} & 
     \includegraphics[width=0.475\textwidth]{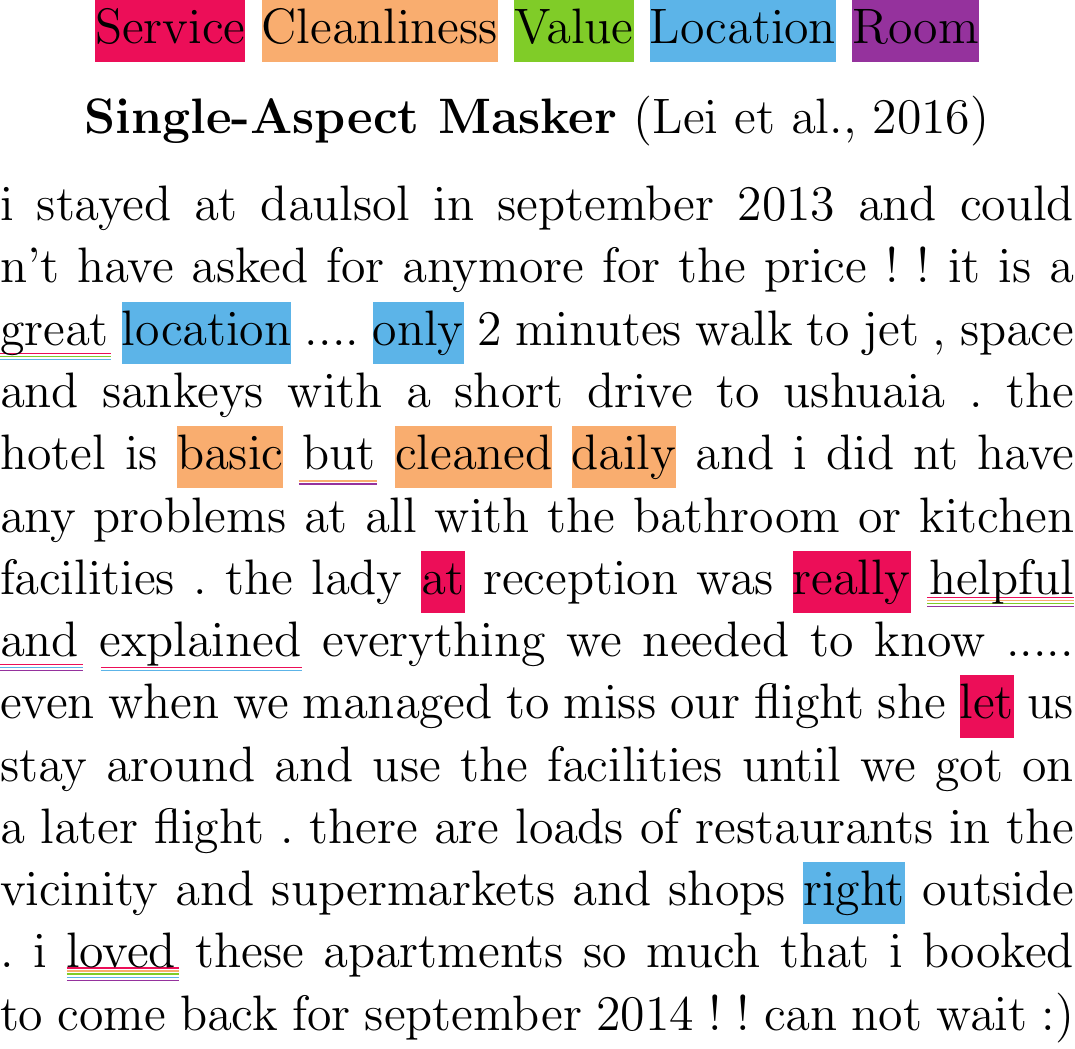} \\\\
     \includegraphics[width=0.475\textwidth]{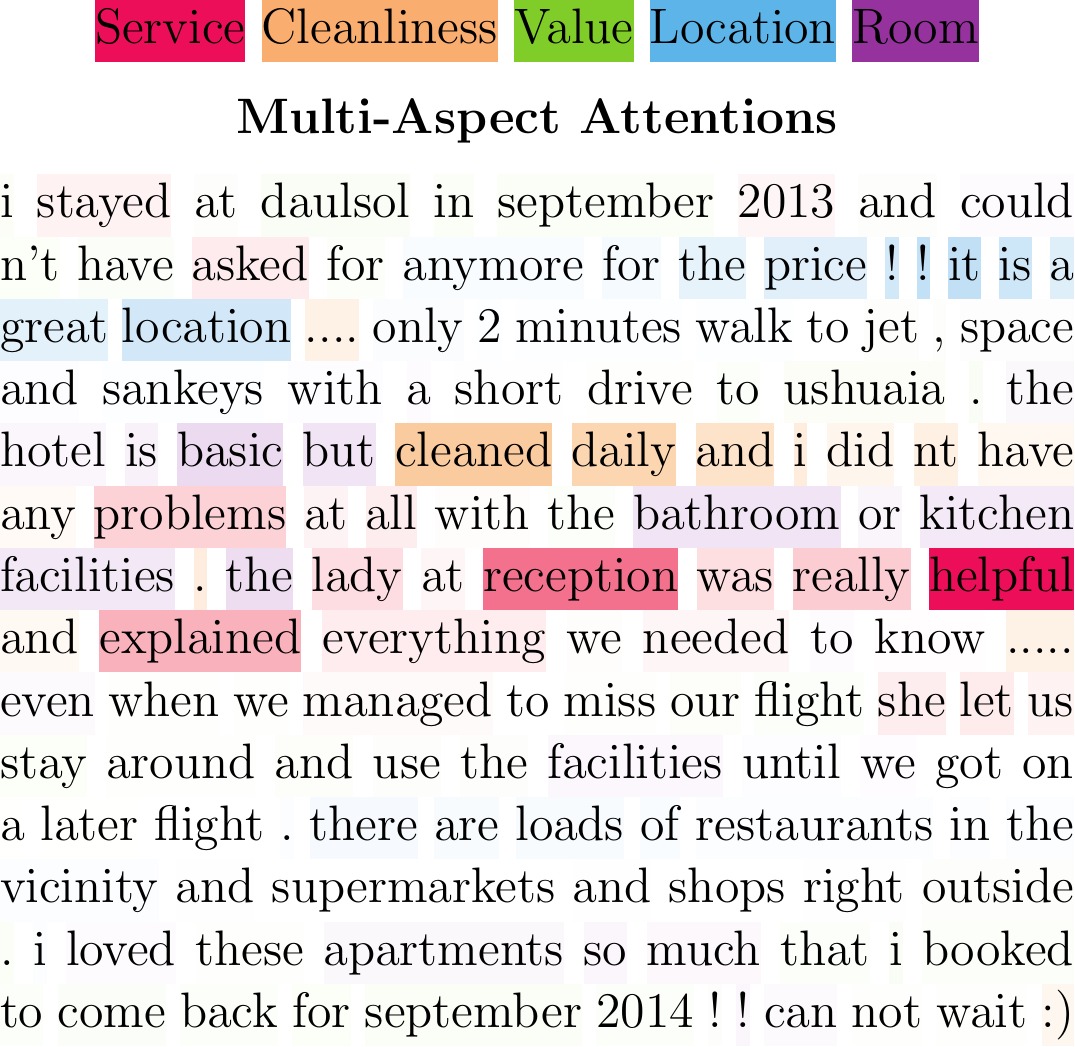} & 
     \includegraphics[width=0.475\textwidth]{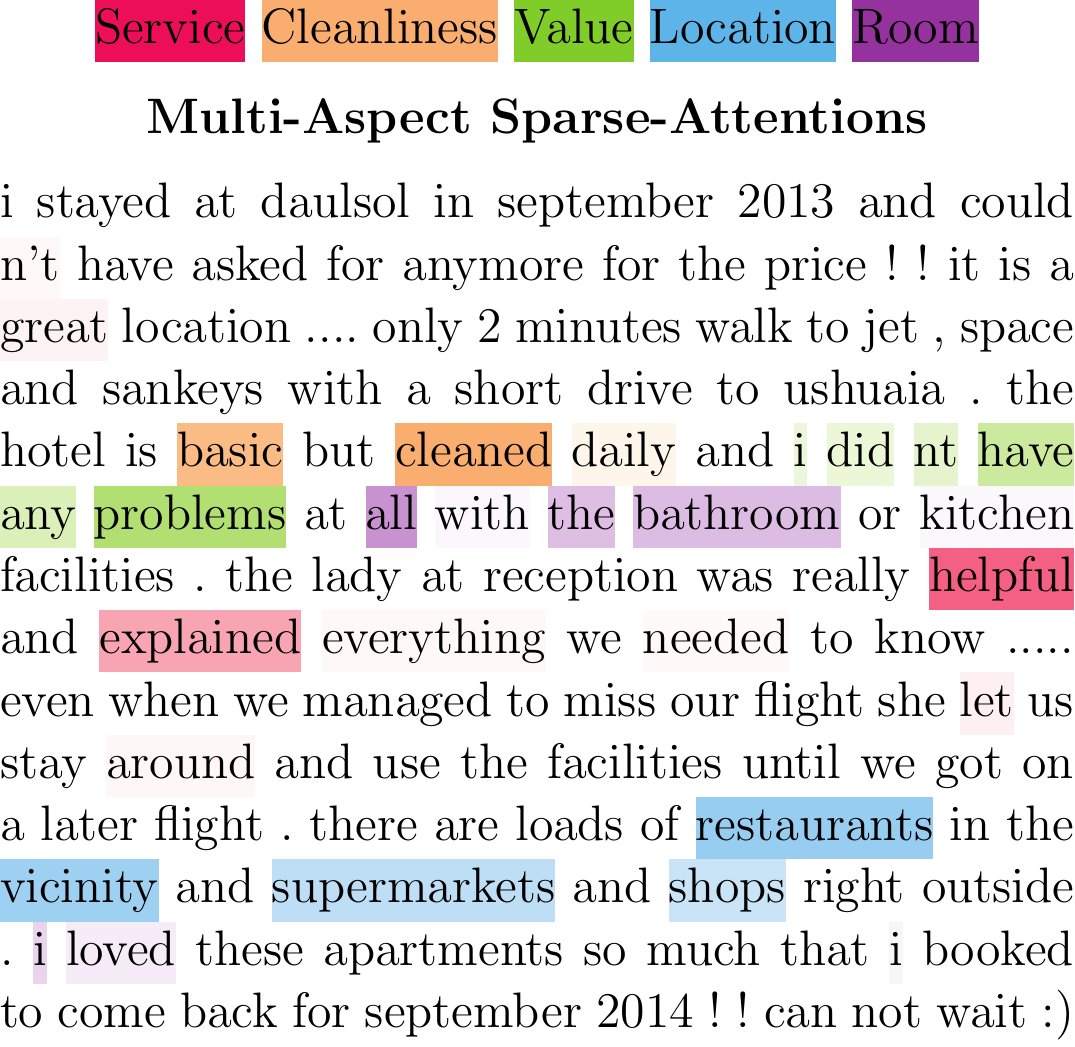}\\\\
\end{tabular}

\caption{\label{sample_hotel_0}A sample review from the \textit{Hotel} dataset, with computed masks from different methods. \textit{MTM} emphasizes consecutive words, identifies essential spans while having a small amount of noise. \textit{SAM} focuses on certain specific words and spans, but labels are ambiguous. The \textit{MAA} model highlights many words, ignores a few crucial key-phrases, but labels are noisy when the confidence is low. \textit{MASA} provides noisier tags than \textit{MAA}.}
\end{figure*}

\begin{figure*}[!htb]
\centering

\begin{tabular}{@{}c@{\makebox[0.25cm]{ }}c@{}}
     \includegraphics[width=0.475\textwidth,height=10cm]{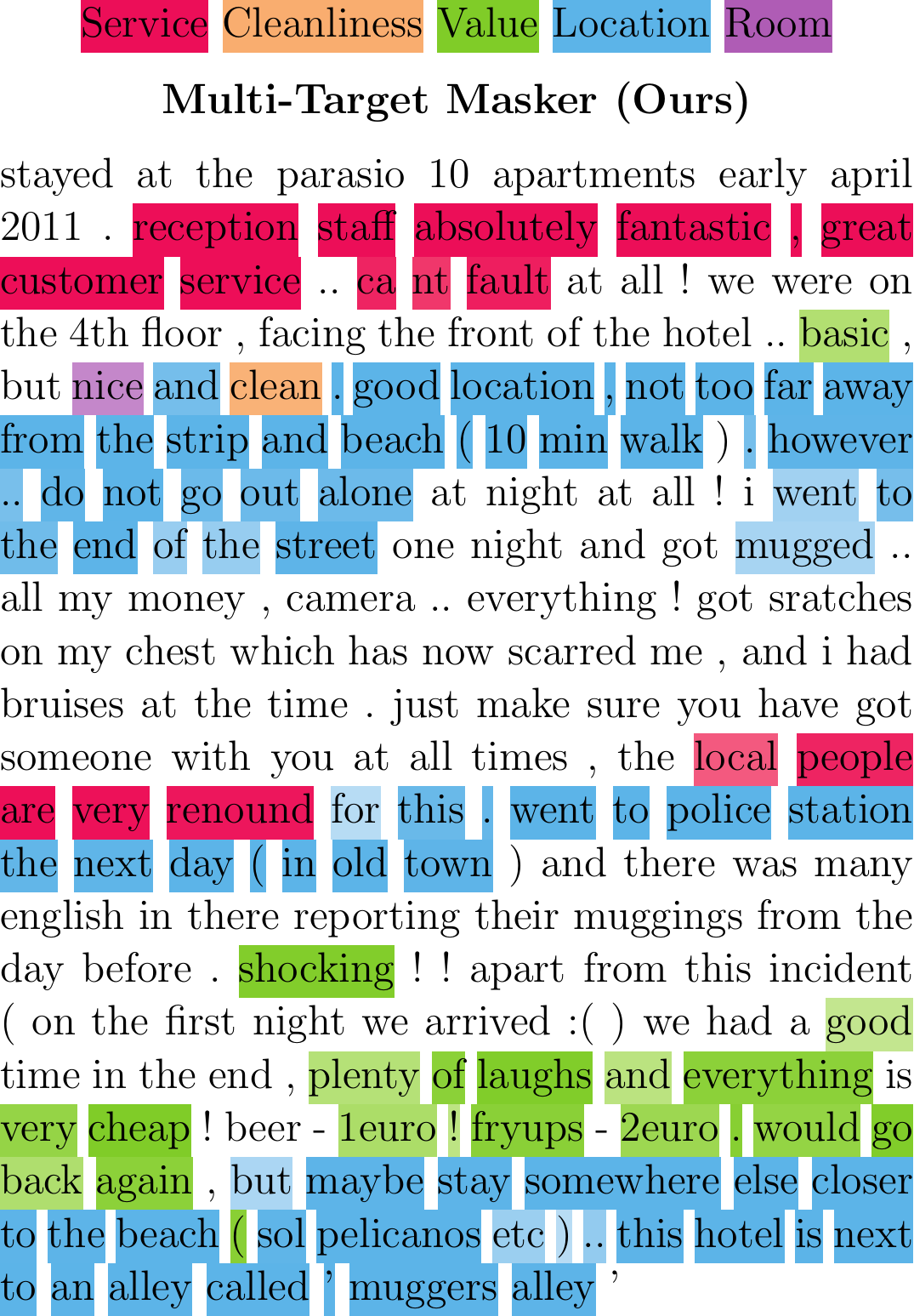} &
     \includegraphics[width=0.475\textwidth,height=10cm]{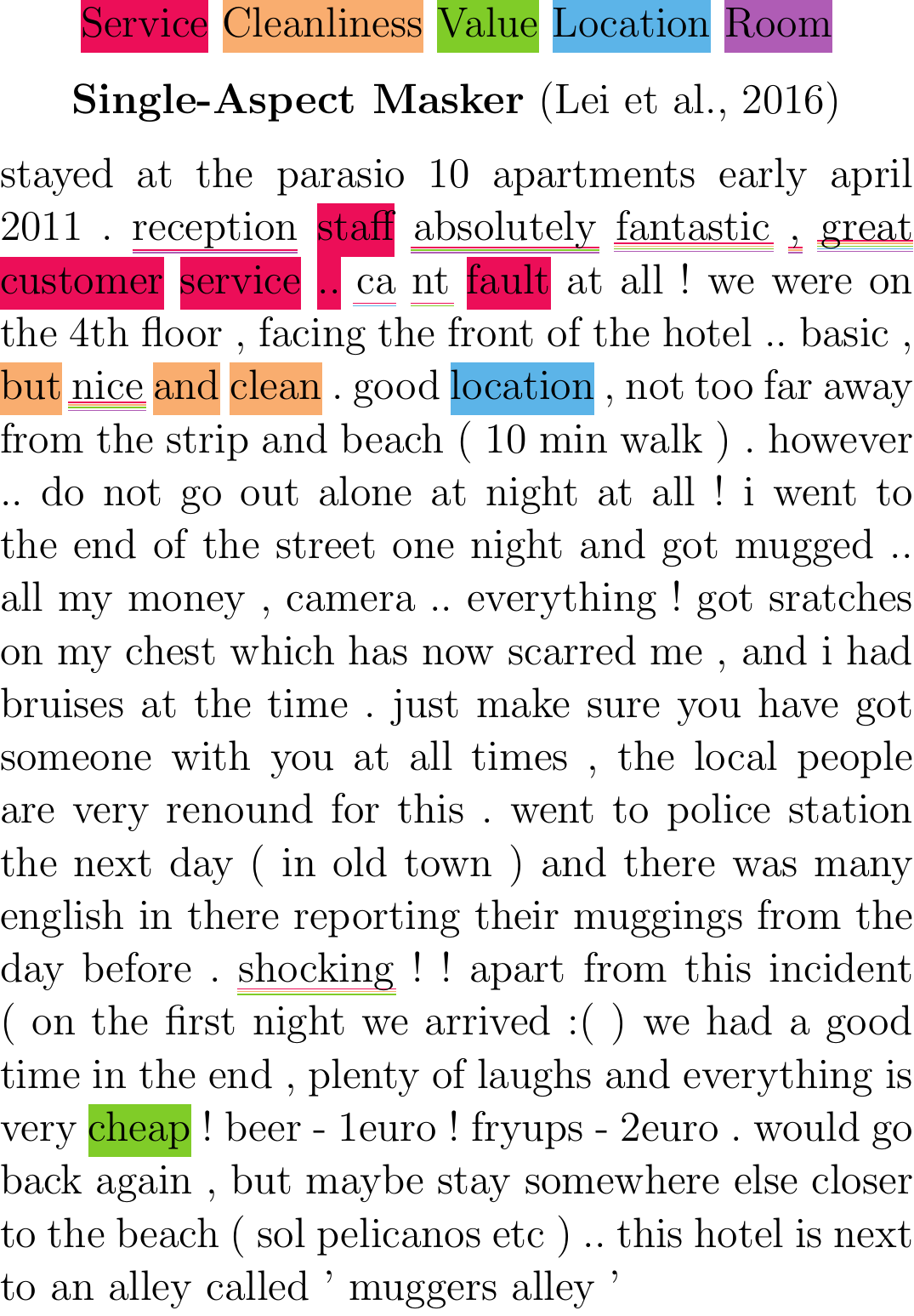} \\\\
     \includegraphics[width=0.475\textwidth,height=10cm]{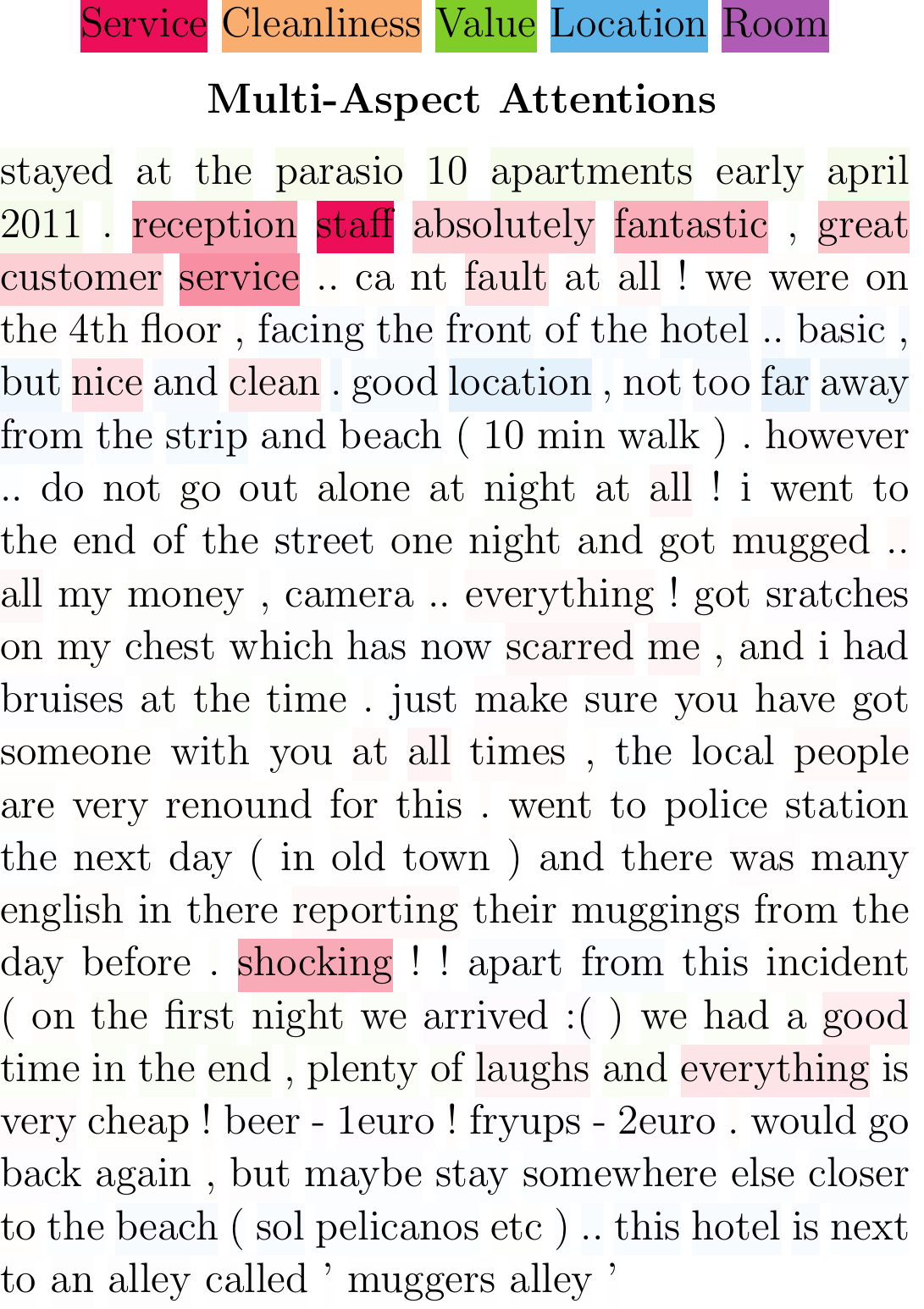} &
     \includegraphics[width=0.475\textwidth,height=10cm]{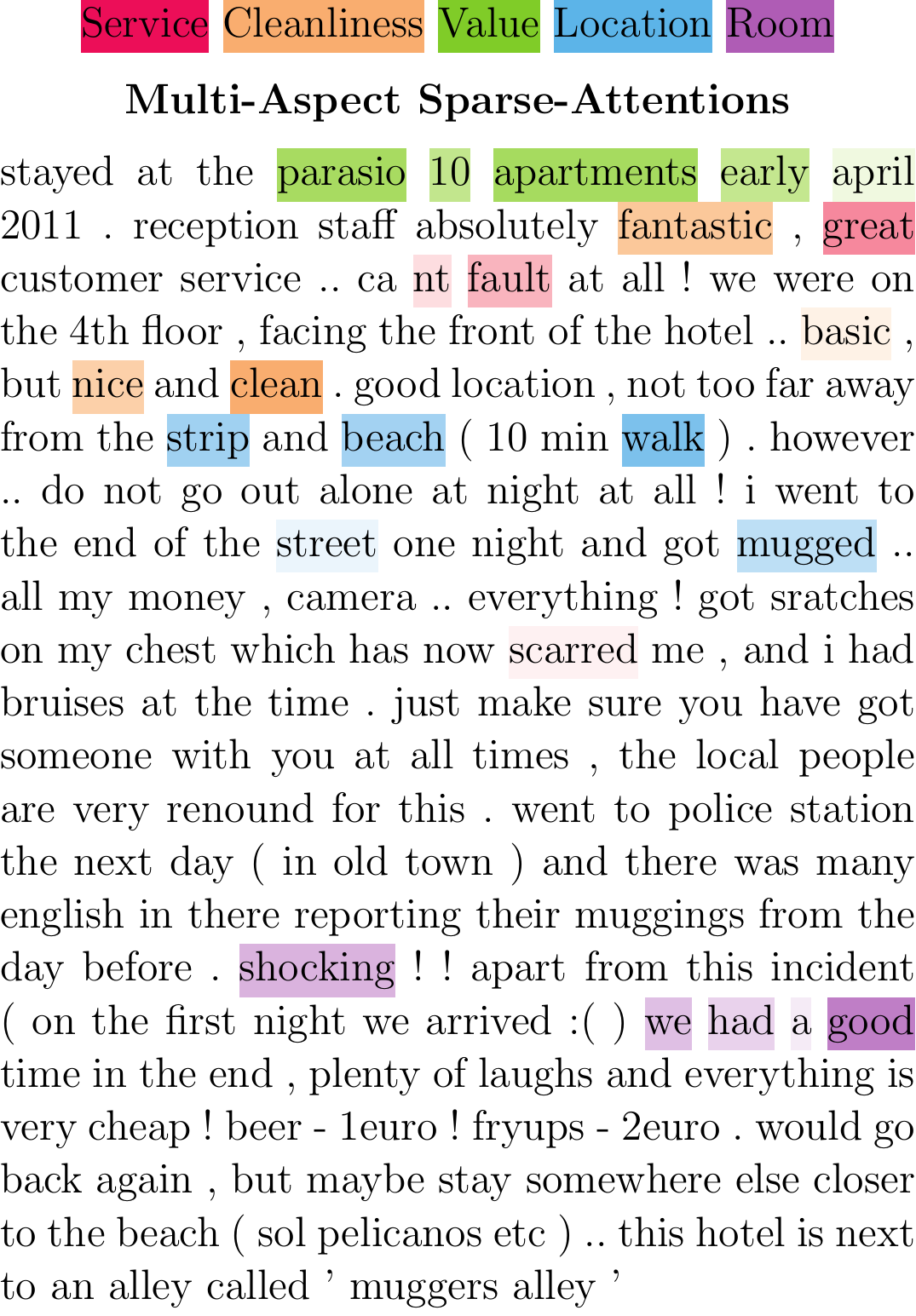}\\\\
\end{tabular}

\caption{\label{sample_hotel_1}A sample review from the \textit{Hotel} dataset, with computed masks from different methods. Our \textit{MTM} model finds most of the crucial span of words with a small amount of noise. \textit{SAM} lacks coverage but identifies words where half are correctly tags and the others ambiguous. \textit{MAA} partially correctly highlights words for the aspects \textit{Service}, \textit{Location}, and \textit{Value} while missing out on the aspect \textit{Cleanliness}. \textit{MASA} confidently finds a few important words.}
\end{figure*}

We randomly sampled reviews from each dataset and computed the masks and attentions of four models: our Multi-Target Masker (\textit{MTM}), the Single-Aspect Masker (\textit{SAM}) \cite{lei-etal-2016-rationalizing}, and two attention models with additive and sparse attention, called Multi-Aspect Attentions (\textit{MAA}) and Multi-Aspect Sparse-Attentions (\textit{MASA}), respectively (see Section~\ref{sec_baselines}). Each color represents an aspect and the shade its confidence. All models generate soft attentions or masks besides SAM, which does hard masking. Samples for the \textit{Beer} and \textit{Hotel} datasets are shown in Figure~\ref{sample_full_beer_0}, \ref{sample_full_beer_1}, \ref{sample_hotel_0}, and \ref{sample_hotel_1}, respectively.

\label{full_beer_sample}

\subsection{Baseline Architectures}
\label{baseline_architecture}

\begin{figure}[!htb]
\centering
\includegraphics[width=.5\textwidth]{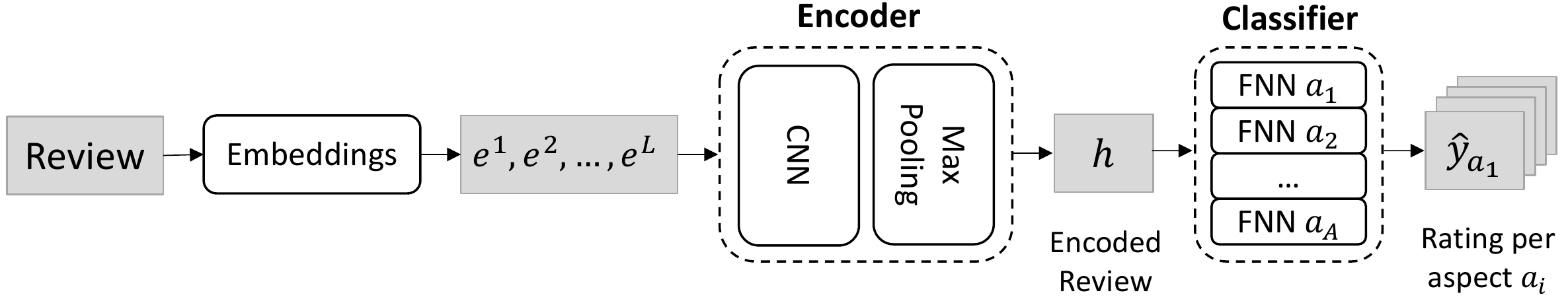} 

\caption{Baseline model $\text{Emb}$ + $\text{Enc}_\textsubscript{CNN}$ + \text{Clf} (\textit{BASE}).}
\end{figure}

\begin{figure}[!htb]
\centering
\includegraphics[width=.5\textwidth]{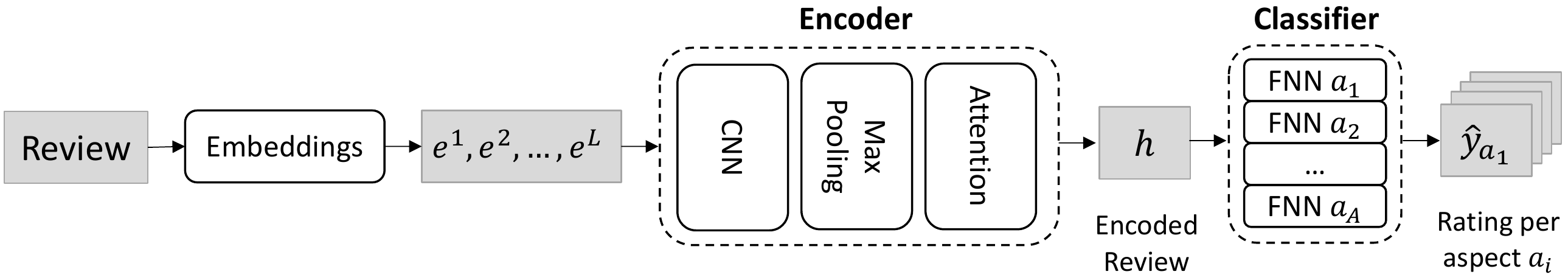}

\caption{Baseline model $\text{Emb}$ + $\text{Enc}_\textsubscript{CNN}$ + $\text{A}_\textsubscript{Shared}$ + \text{Clf} (\textit{SAA}, CNN variant).}
\end{figure}

\begin{figure}[!htb]
\centering
\includegraphics[width=.5\textwidth]{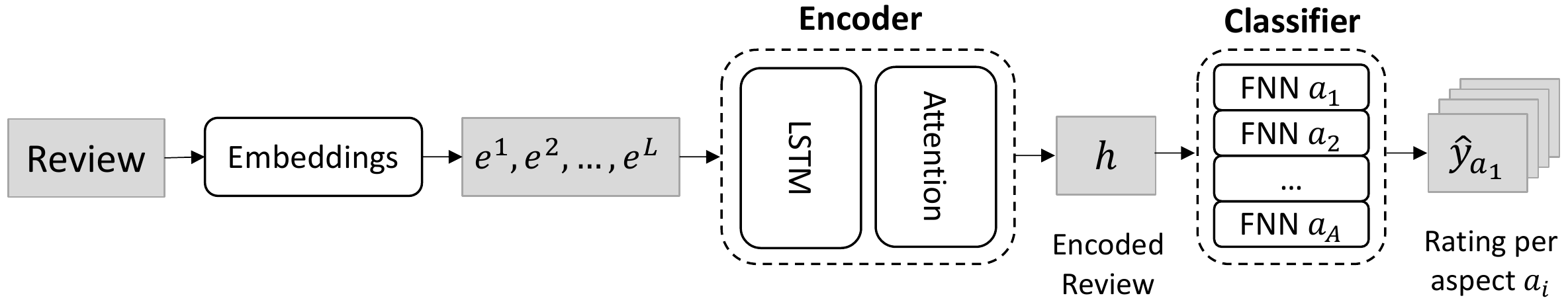}

\caption{Baseline model $\text{Emb}$ + $\text{Enc}_\textsubscript{LSTM}$ + $\text{A}_\textsubscript{Shared}$ + \text{Clf} (\textit{SAA}, LSTM variant).}
\end{figure}

\begin{figure}[!htb]
\centering
\includegraphics[width=.5\textwidth]{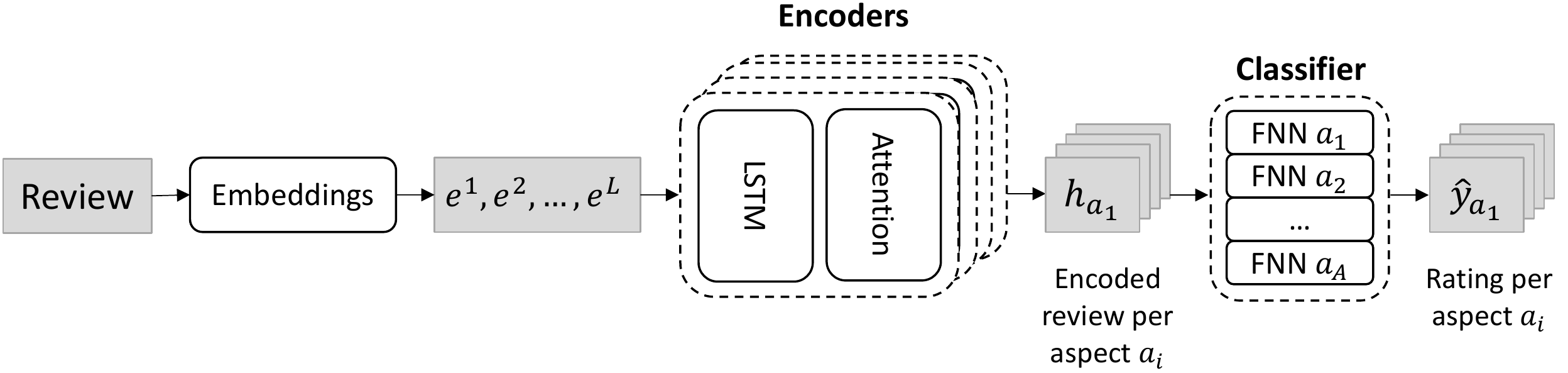} 

\caption{Baselines $\text{Emb}$ + $\text{Enc}_\textsubscript{LSTM}$ + $\text{A}^\textsubscript{[Sparse]}_\textsubscript{Aspect-wise}$ + \text{Clf}. Attention is either additive (\textit{MAA}) or sparse (\textit{MASA}).}
\end{figure}

\section{Additional Training Details}
\label{app:additional_training}
Most of the time, the model converges under $20$ epochs (maximum of $20$ and $3$ minutes per epoch for the \textit{Beer} and \textit{Hotel} dataset, respectively. The range of hyperparameters are the following for MTM (similar for other models).
\begin{itemize}
	\item Learning rate: $[0.001, 0.0005, 0.00075]$;
	\item Hidden size: $[50, 100, 200]$;
	\item Filter numbers (CNN): $[50, 100, 200]$;
	\item Bi-directional (LSTM): $[True, False]$;
	\item Dropout: $[0, 0.1, 0.2]$;
	\item Weight decay: $[0, 1e^{-6}, 1e^{-8}, 1e^{-10}]$;
	\item Gumbel Temperature $\tau$ :$[0.5, 0.8, 1.0, 1.2]$;
	\item $\lambda_{sel}$: $[0.01, 0.02, 0.03, 0.04, 0.05]$;
	\item $\lambda_p$: $[0.05, 0.06, 0.07, 0.08, 0.09, 0.1, 0.11, 0.12, 0.13, 0.14, 0.15]$;
	\item $\lambda_{cont}$: $[0.02, 0.04, 0.06, 0.08, 0.10]$;
\end{itemize}
We used a computer with the following configuration: 2x Intel Xeon E5-2680, 256GB RAM, 1x Nvidia Titan X, Ubuntu 18.04, Python 3.6, PyTorch 1.3.1, CUDA 9.2.

\section{Applying MTM for Image Classification}

Our method can be applied to different types of inputs like images. In that case, the words in the text data are replaced~by regions or patches in an image, obtained by a classical pre-trained convolutional neural network. The sparsity constraint stays similar, and the continuity is based on 2D distances instead. We will investigate this type of data in future work.

\section{Impact of the Length of the Rationales}
\label{app:impact_length}

We are interested in studying \begin{enumerate*}
 \item the distribution of the rationales' lengths, and
 \item the effect of the length in terms of the precision, recall, and F1 score.
 \end{enumerate*} In Section~\ref{subsub_prec}, we report only the precision on three aspects at a similar number of selected words, for a fair comparison with prior work, on the human sentence-level aspect annotations. 
 
We use the \textit{Beer} dataset and also consider the extra aspect \textit{Overall} (which leads to five aspects in total), and we train MTM on it as in Section~\ref{experimental_details}. We compute the distribution of the explanation length (in percent) on the validation set for each aspect $a_i, i=1,...,5$. We calculate all the $100$ percentiles $P^{a_i}_p$. Then, we infer the sub-masks $M_{a_i}$ and generate the rationale by selecting each word $x^\ell$, whose relevance towards $a_i$ satisfies $P(m_{a_i}^\ell|x^\ell) \ge P^{a_i}_p$. Finally, we compute the precision, recall, and F1 score on the human annotation for each percentile~$P^{a_i}_p$.

\begin{figure}[!t]
\centering
\includegraphics[width=0.45\textwidth]{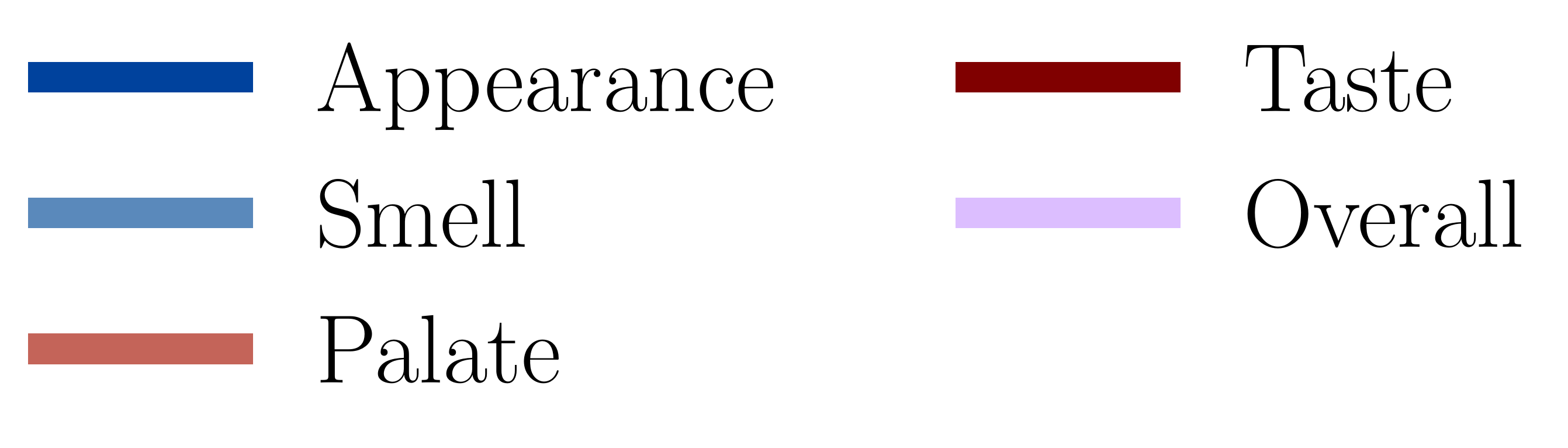}\\
\includegraphics[width=0.35\textwidth]{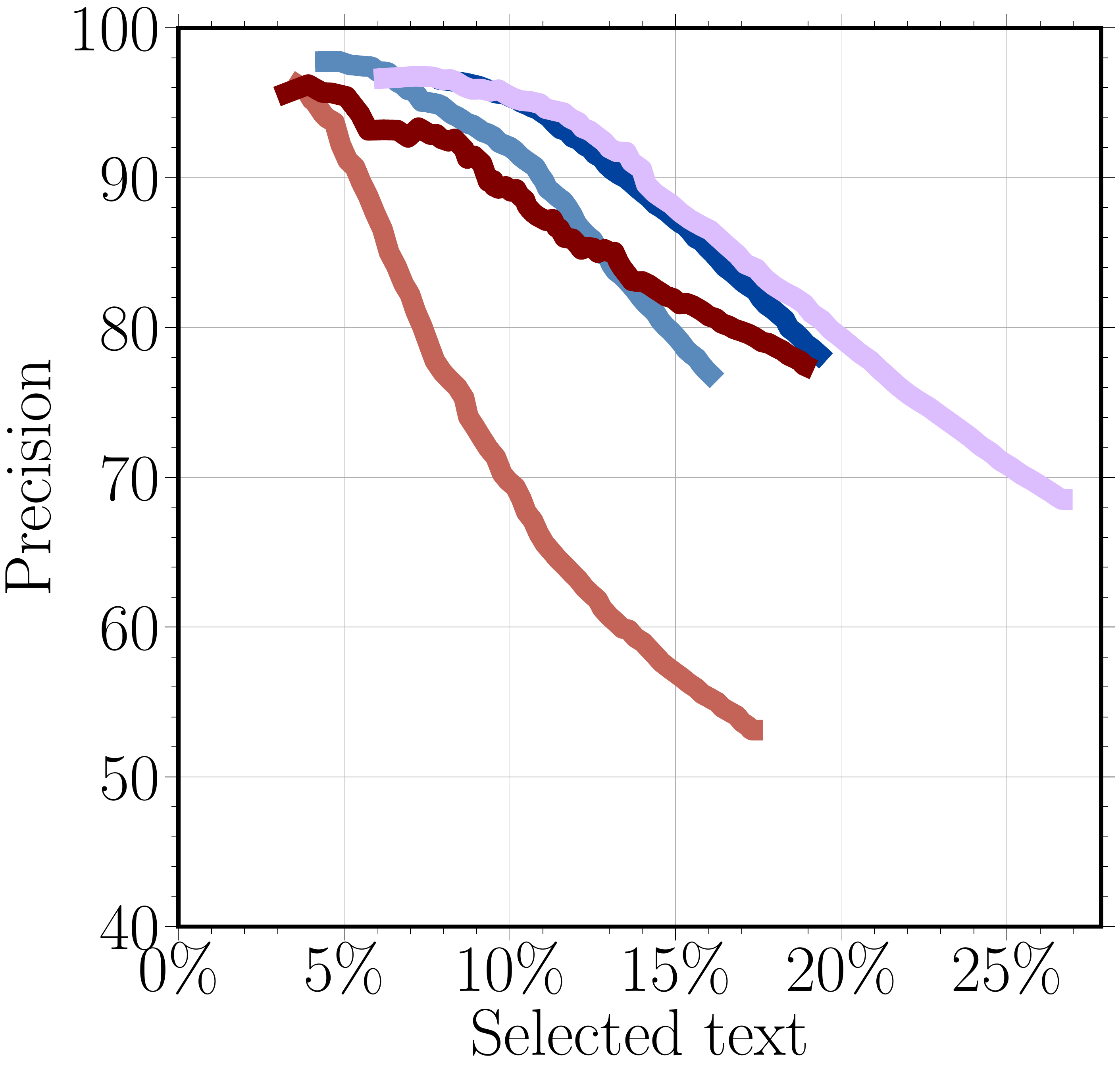}\\
\includegraphics[width=0.35\textwidth]{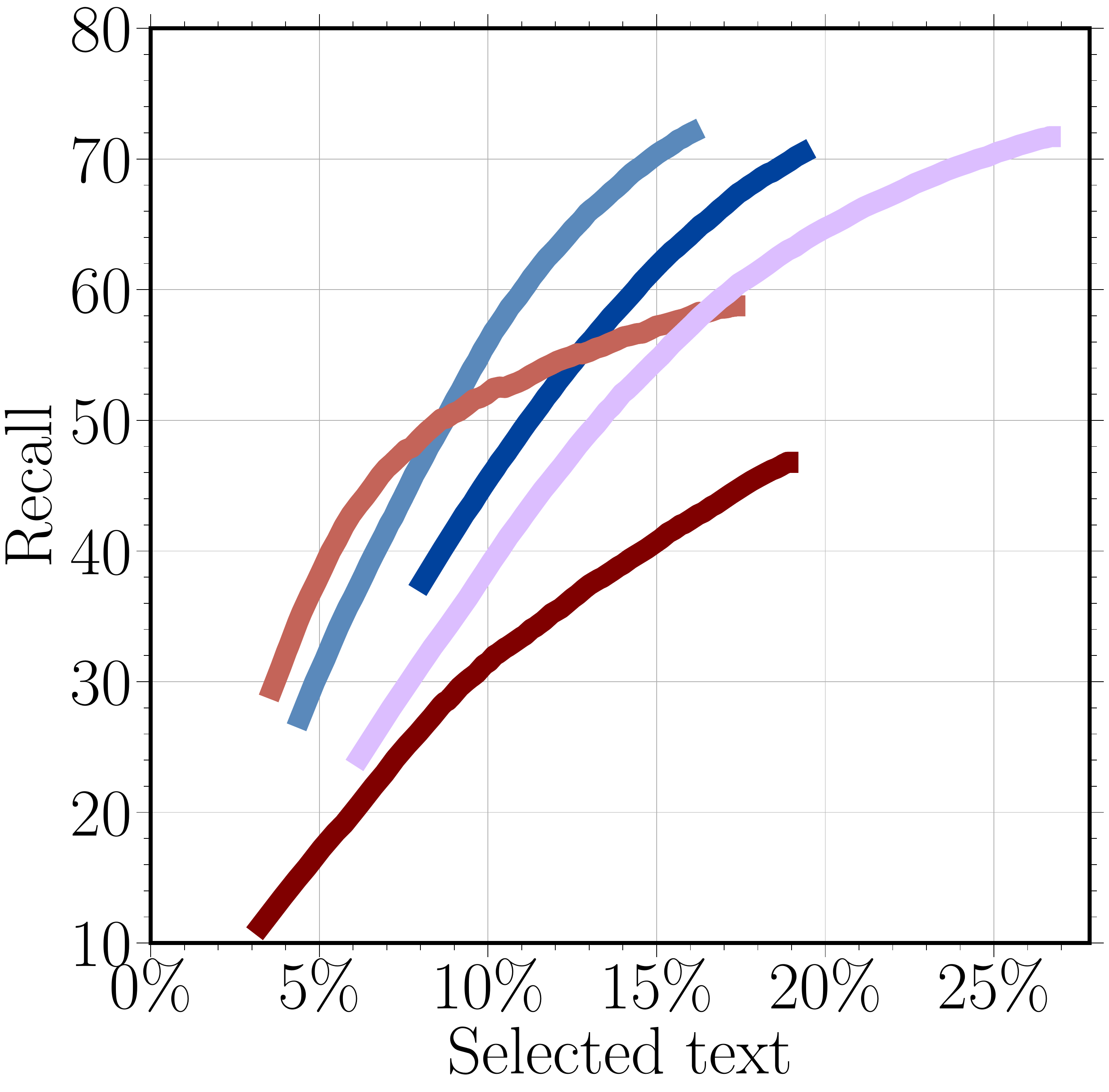}\\
\includegraphics[width=0.35\textwidth]{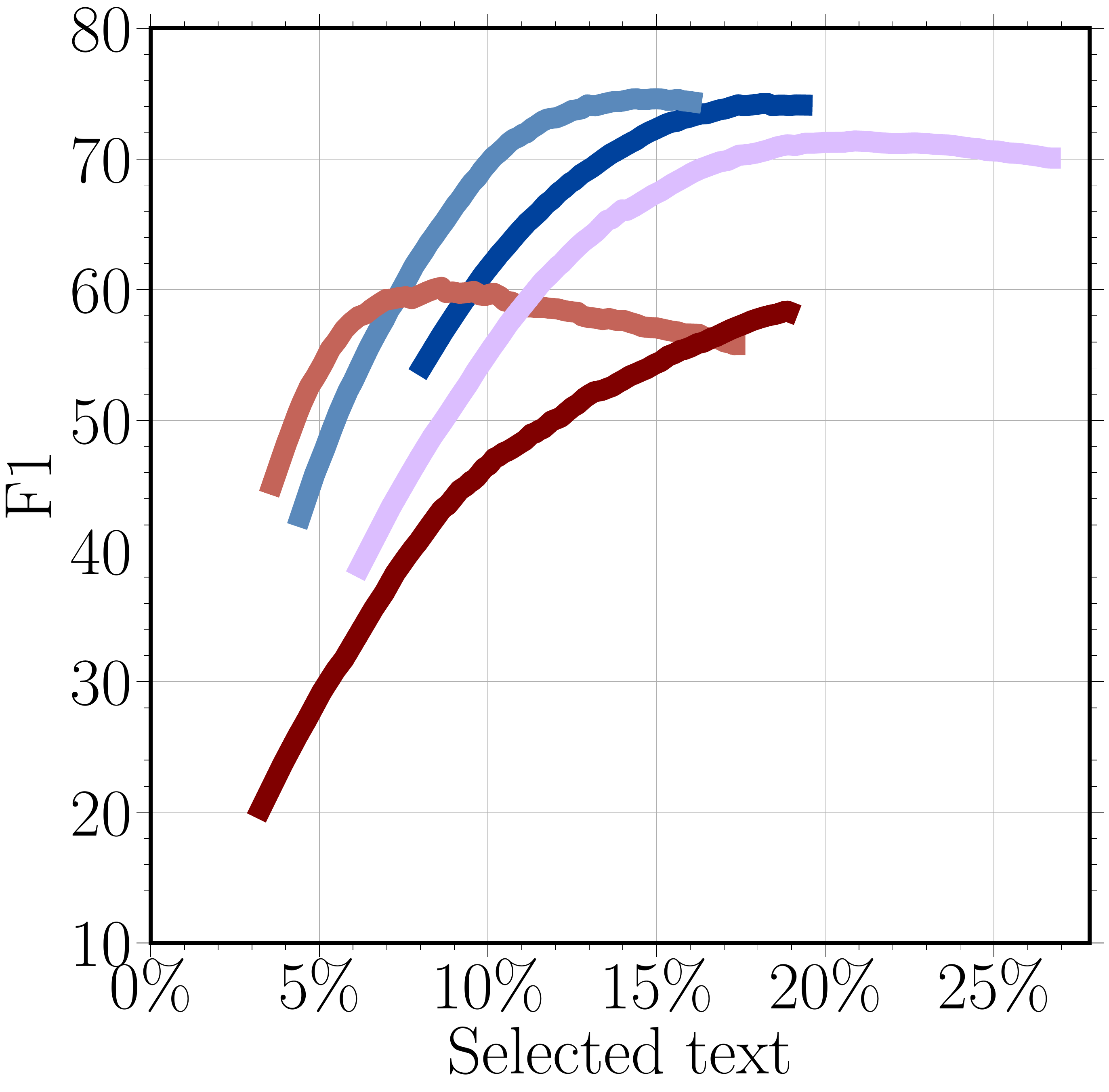}
\caption{\label{app_he_ann}Performance related to human evaluation, showing the precision, recall, and F1 scores of the rationales for each aspect (with the overall rating) of the \textit{Beer} dataset. The percentage of words indicates the number of highlighted words of the full~review.}
\end{figure}

Figure~\ref{app_he_ann} shows the result of the five aspects. First, we observe that the length distribution of the rationales for the aspect \textit{Overall} is the most spread compared to the other aspects, and these of the aspect \textit{Smell} the least; the three left aspects share a similar distribution.

In terms of precision, we notice that the aspect \textit{Palate} drops quickly compared to the other ones that decrease linearly when we augment the portion of selected text. According to RateBeer\footnote{url{https://www.ratebeer.com/Story.asp?StoryID=103}}, the aspect \textit{Palate} is the most difficult one to rate, which confirms our findings. For most aspects the precision remains high after selecting $75\%$ of the text ($5 \cdot 15\%$), showing the effectiveness of our approach. In terms of recall, they all increase unsurprisingly when highlighted more~words.

Finally, we show that MTM achieves a very high precision~($>90\%$) by highlighting only $25\%$ of the document, that reduces the cognitive load of a user to identify the important parts of the document.

\label{hotel_samples}

\end{document}